\pdfoutput=1

\documentclass[a4paper,fleqn]{cas-dc}

\usepackage[numbers,sort&compress]{natbib}
\usepackage{tikz}
\usepackage{xcolor}
\usepackage{placeins}

\ExplSyntaxOn
\NewDocumentCommand{\setcaptionwidth}{m}{ \dim_set:Nn \l_tbl_width_dim {#1} }
\ExplSyntaxOff
\newcommand{\stretchtables}{%
  \renewenvironment{tabular}[1]{\begin{tabular*}{\linewidth}{@{\extracolsep{\fill}}##1}}{\end{tabular*}}}

\begin{document}
\let\WriteBookmarks\relax

\shorttitle{Sensorless Strain-Bounded Grasping of Fruit}
\shortauthors{Y. Shuto and D. Vasconcellos Vargas}

\title[mode=title]{Sensorless strain-bounded grasping of fruit}

\author[1]{Yusei Shuto}[orcid=0009-0007-4232-2779]
\cormark[1]
\ead{syutou.yusei.061@s.kyushu-u.ac.jp}

\author[1,2]{Danilo {Vasconcellos Vargas}}
\ead{vargas@inf.kyushu-u.ac.jp}

\affiliation[1]{organization={Kyushu University},
            city={Fukuoka},
            country={Japan}}

\affiliation[2]{organization={MiraiX},
            city={Fukuoka},
            country={Japan}}

\cortext[1]{Corresponding author}

\begin{abstract}
Robotic post-harvest handling must hold produce without bruising it, yet
fruit stiffness varies several-fold with ripeness, so no fixed grip
force suits every item. We instead bound compression strain using only
the encoder and motor effort of a servo gripper. Given a strain bound
$\varepsilon$ and a stiffness floor $k_{\min}$ set once per crop and
gripper, the controller estimates object size $\hat{D}$ at contact,
refines stiffness $\hat{k}$ conservatively by probing, and stops at
$\hat{F} = \varepsilon \hat{k} \hat{D}$, keeping true strain below
$\varepsilon$ with high probability above a quantified contact-detection
floor. In simulation, where fixed force thresholds either drop heavy
objects or damage soft ones, the proposed controller carries $92\,\%$ of
objects within its operating range with $0.3\,\%$ damage. On cherry and
medium tomatoes, setting $\varepsilon$ above the detection floor scaled
the stop force threefold across fruit sizes, carrying all fruits while
reducing markings on cherry tomatoes from $8/10$ to $1/10$; on medium
tomatoes both marked $3/6$.
\end{abstract}

\begin{keywords}
Strain-bounded grasping \sep online stiffness estimation \sep
sensorless force control \sep proprioceptive sensing \sep
deformable object manipulation \sep post-harvest handling
\end{keywords}

\maketitle

\begin{figure}[pos=t]
  \centering
  \includegraphics[width=\columnwidth]{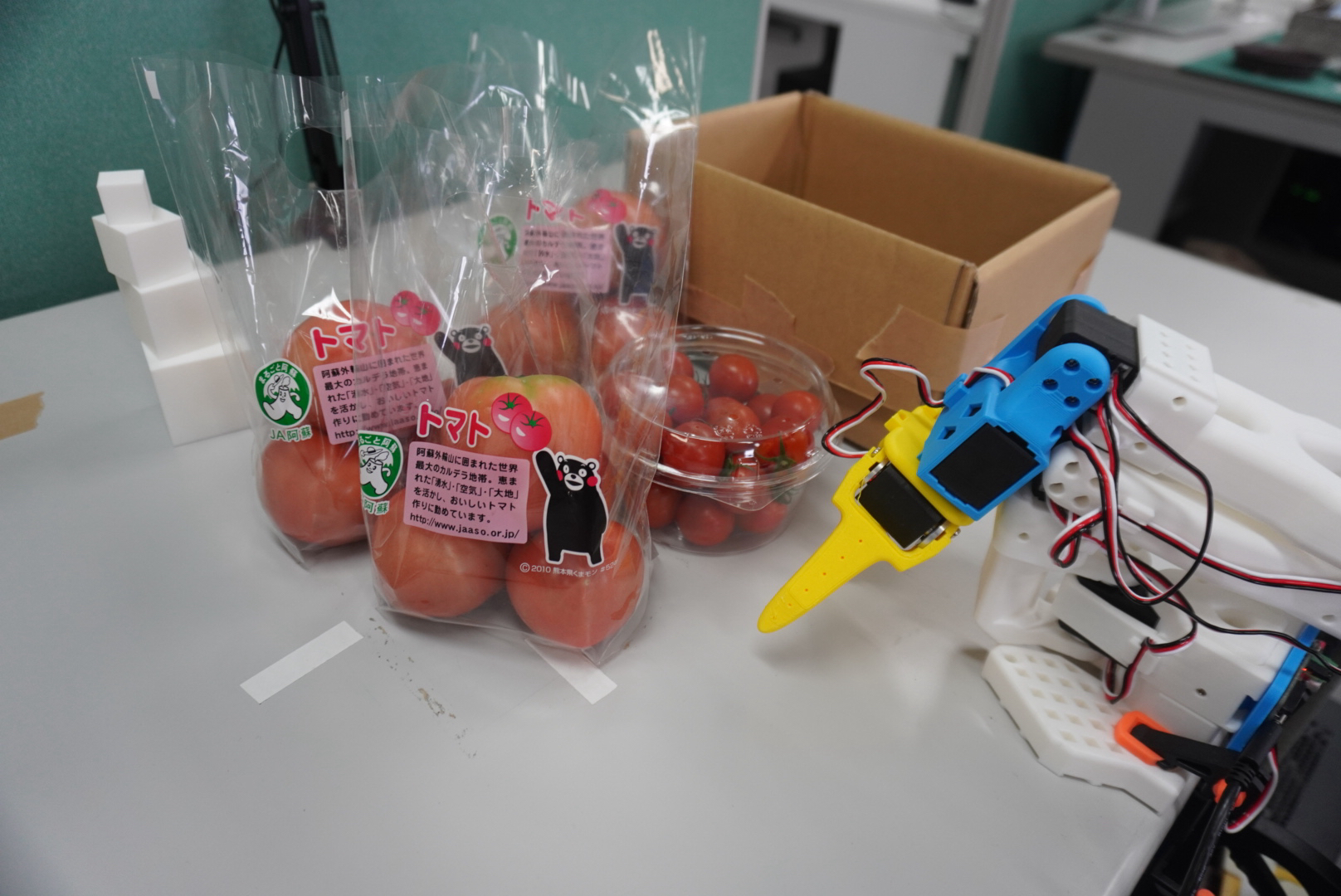}
  \caption{The SO-ARM101 gripper, the tomatoes and the box of the
    hardware trials. The controller reads only the jaw position $x$
    and the motor effort $I$ of the gripper servo, and the operator
    sets only the strain bound $\varepsilon$ and the stiffness floor
    $k_{\min}$.}
  \label{fig:teaser}
\end{figure}

\section{Introduction}
\label{sec:intro}

Robotic post-harvest handling of fruit needs a grasp that holds the
fruit through transport without damaging
it~\cite{chauhan2022robotics, wang2025towards}. Fruit of one
species varies too much for a fixed grip force. The compression
stiffness of whole tomatoes falls from about $10\,000\,\mathrm{N/m}$
when firm to about $2000\,\mathrm{N/m}$ when
overripe~\cite{sirisomboon2012evaluation}, so a force that holds the
heaviest fruit crushes the softest, and a force safe for the softest
drops the heaviest. The grip force has to adapt to each
fruit~\cite{wang2026adaptive, yu2025grasping}.

Damage follows the compression strain of the fruit rather than the
force~\cite{li2017mathematical, li2013internal}. A force $F$ compresses
a fruit of stiffness $k$ and size $D$ by the strain $F/(kD)$, so
bounding the strain needs both. Tactile skins such as
GelSight~\cite{yuan2017gelsight} and vision--tactile
feedback~\cite{han2024learning} can sense the contact, but they add
cost, fragility and wiring that a handling robot can hardly afford. The
encoder and motor effort that every actuated gripper already provides
give a force estimate~\cite{giannico2017evaluation,
ballesteros2020proprioceptive}, and the jaw gap at contact gives the
size. The stiffness, however, can only be learned by pressing the
fruit, which is exactly what has to be limited.

We learn the stiffness within the bound. The first loading assumes the
softest fruit of the crop and stops where that fruit would reach the
strain bound $\varepsilon$. Each loading refines the stiffness
estimate, which is kept on its low side so that each stop stays within
the bound (\S\ref{sec:alpha}) and the next loading can press a little further. The
operator sets only $\varepsilon$ and the stiffness of the softest
fruit, once per crop and gripper, whereas a fixed force must be
retuned for each size and ripeness. Whole-fruit tests report
both~\cite{li2017mathematical, sirisomboon2012evaluation}, but at the
jaw contact the published strain bound is conservative while the
published stiffness is too high, so the floor must be set there
(\S\ref{sec:future}).

One cost cannot be avoided. Without a force sensor, contact is
detected only once the motor effort rises above its noise, and by then
the fruit is already compressed. This detection floor limits how small
and soft a fruit the method can handle at a given loading speed, and
we quantify it. Fig.~\ref{fig:teaser} shows the gripper and the
tomatoes of the hardware trials, which use only the servo's own
signals.

\paragraph{Contributions.}
\begin{itemize}
  \item A strain-bounded grasp controller that uses only the encoder
    and the motor effort and estimates the stiffness by probing within
    the bound.
  \item A conservative strain bound under stated assumptions
    (\S\ref{sec:alpha}).
  \item An operating range set by the contact-detection floor and by
    the force the bound allows.
  \item Validation in simulation against fixed-force and size-scaled
    stops, and on tomatoes against fixed-force grasps.
\end{itemize}

\section{Related Work}
\label{sec:related}

\subsection{Robotic fruit handling}
Labor shortages drive robots into the post-harvest handling of fresh
produce~\cite{chauhan2022robotics}. Fruit is easily bruised in
handling and loses much of its market value once
damaged~\cite{opara2014bruise, al2022mechanical}, so the grasp
matters. End effectors developed for fruit range from rigid grippers
and cutting or suction tools for apple~\cite{zhang2024automated},
strawberry~\cite{xiong2020autonomous} and sweet
pepper~\cite{lehnert2017autonomous} to soft and pneumatic grippers that
protect the fruit by their compliance~\cite{wang2023soft,
wang2025towards}; most soft crop grippers are tested on pick-and-place
without detachment from the plant~\cite{elfferich2022soft}. Recent grippers add tactile and slip sensing to
adjust the force online~\cite{liu2024soft}. The variety of fruit in
size, firmness and ripeness still makes any preset grip
unreliable~\cite{wang2025towards}.

\subsection{Tactile and vision sensing for deformable grasping}
Closest to our task is Han \emph{et al.}~\cite{han2024learning}, who
combine a GelSight tactile camera~\cite{yuan2017gelsight} with RGB
images in a transformer to classify each grasp of a deformable object
as safe, slipping or damaging. The output is a class label, not a
strain bound the user can set, and success varies with shape, from
$38\,\%$ on bananas to $90\,\%$ on round fruit. Other visual--tactile
methods also assess the grasp state of deformable
objects~\cite{cui2020grasp}. All of them need dedicated tactile
hardware, whereas we use only the encoder and motor effort that every
servo gripper already provides.

\subsection{Sensorless force estimation from motor current}
A robot can estimate contact force from its own actuators without a
force sensor. Classical methods detect contact and recover the
external force from the generalized momentum and motor
current~\cite{deluca2006collision}, and later work refines the
estimate with disturbance observers and learned
friction~\cite{liu2021sensorless} or deep temporal
networks~\cite{wu2025contact}. In grasping, the force--current relation
of an underactuated gripper can be
calibrated~\cite{giannico2017evaluation}, and finger proprioception can
estimate contact force without a tactile
skin~\cite{ballesteros2020proprioceptive}. Sensor-free methods for
deformable objects predict grasp poses or deformation
properties~\cite{yu2023defgrasp}.

None of these methods turns the estimate into a compression strain
with a bound the user sets, on objects of unknown stiffness and
without calibration. Estimating that stiffness online from small
compressions is hard, and is usually done by deliberate probing or
learned estimation with uncertainty~\cite{kutsuzawa2024learning}. We
probe within the bound and keep the estimate on its conservative side
(\S\ref{sec:alpha}). Sensorless contact detection on soft objects also
fires only after the object is already compressed, and to our
knowledge prior work does not measure this cost. Our floor analysis
(\S\ref{sec:floor}) quantifies it and links it to the loading speed.

\section{Method}
\label{sec:method}

\subsection{Overview}
\begin{figure*}[pos=tp]
  \centering
  \includegraphics[width=\textwidth]{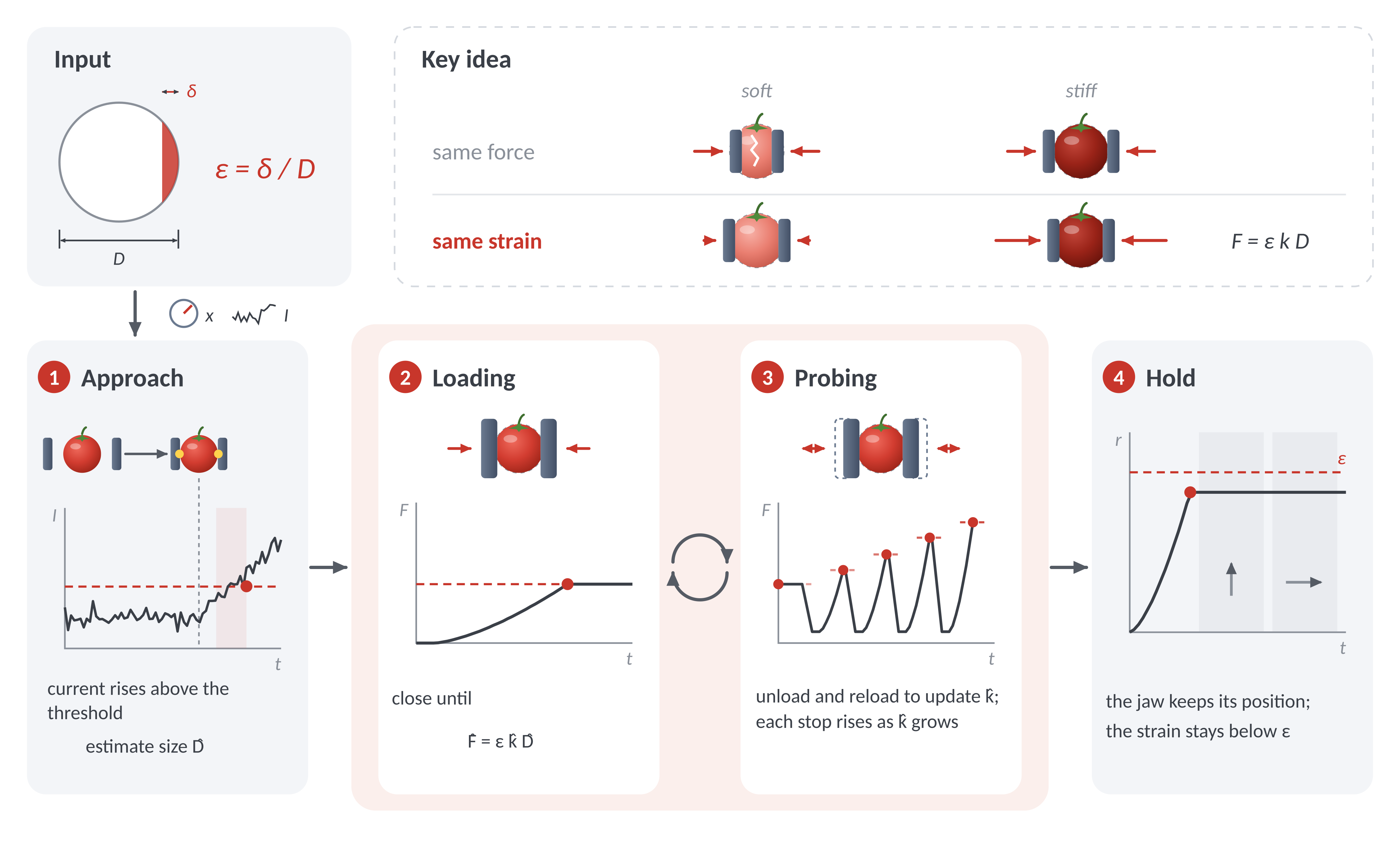}
  \caption{Method overview. The operator sets the strain bound
    $\varepsilon = \delta/D$, the compression the object may take over
    its size. The key idea: one force bruises a soft fruit and lets a
    stiff one slip, while one strain needs a force that follows the
    stiffness, $F = \varepsilon k D$. The controller reads only the jaw
    position $x$ and the motor current $I$. (1)~Approach: the jaw closes
    freely until contact makes the current rise above a threshold,
    which gives the size estimate $\hat{D}$. (2)~Loading: the jaw closes
    until the force proxy reaches $\varepsilon\hat{k}\hat{D}$, with
    $\hat{k}$ starting at the floor $k_{\min}$. (3)~Probing: each step
    updates $\hat{k}$ and raises the stop force, either by cyclic
    unload--reload or by monotonic stepping along one closing pass
    (\S\ref{sec:probe}). (4)~Hold: the jaw keeps its position through the
    lift and the transport.}
  \label{fig:pipeline}
\end{figure*}

The controller reads two scalar signals at $50\,\mathrm{Hz}$: the jaw
position $x(t)$ from the servo encoder, and a motor-effort signal
$I(t)$. It commands the jaw velocity through a single normalized action
$a(t) \in [-1, 1]$. From these two signals alone, with no tactile
sensor and no force-torque measurement, it estimates the object's
compression strain $\hat{r}(t)$ and stops the jaw when $\hat{r}$
reaches the user-specified bound~$\varepsilon$.
Each grasp passes through four phases (Fig.~\ref{fig:pipeline}).
\begin{enumerate}
  \item \emph{Approach.} The jaw closes until contact is detected, and
    the object size is estimated (\S\ref{sec:contact}).
  \item \emph{Loading.} The jaw closes slowly while the controller
    estimates the object's stiffness (\S\ref{sec:online-k}).
  \item \emph{Probing.} The jaw steps the stop force up from the floor as
    the stiffness estimate improves (\S\ref{sec:probe}).
  \item \emph{Hold.} The jaw holds its position through the lift
    (\S\ref{sec:stop-policy}).
\end{enumerate}
The operator sets the strain bound $\varepsilon$ and a stiffness floor
$k_{\min}$ once per crop and gripper. Two quantities are calibrated once per gripper
(three on hardware, \S\ref{sec:hardware}),
and nothing is measured per fruit. \S\ref{sec:alpha} states the
conditions under which the true strain stays below $\varepsilon$.

\subsection{Force and strain estimates}
\label{sec:stop}

In simulation the effort signal $I$ is the motor current; on the
Feetech servo it is the PWM-derived load register. Following
proprioceptive force estimation~\cite{giannico2017evaluation,
deluca2006collision}, we estimate the contact force as
\[
  \hat{F}(t) \;=\; \max\!\bigl(0,\, (I(t) - I_0)/\alpha\bigr),
\]
where $I_0$ is the pre-contact effort baseline (\S\ref{sec:contact})
and $\alpha$ is the effort-per-force coefficient ($\mathrm{A/N}$ in
simulation, load units$/\mathrm{N}$ on the servo), calibrated once per
gripper. For control, $\hat{F}$ is smoothed by an exponential moving
average ($\beta = 0.3$). Any calibration error in $\alpha$ cancels
whenever $\hat{k}$ is estimated from $\hat{F}$ (\S\ref{sec:alpha}).

The controlled quantity is the compression strain, the depth $\delta$
by which the jaws compress the object below its undeformed size $D$:
\[
  r \;=\; \frac{\delta}{D},
\]
quoted together with $\varepsilon$ as a percentage. Using $\delta = F/k$,
the controller estimates strain as
\[
  \hat{r}(t) \;=\; \frac{\hat{F}(t)}{\hat{k}\,\hat{D}},
\]
from the size estimate $\hat{D}$ (\S\ref{sec:contact}) and the
stiffness estimate $\hat{k}$ (\S\ref{sec:online-k}).

\subsection{Approach}
\label{sec:contact}
\label{sec:size}

Let $I_0$ be the motor-effort baseline, taken as the mean of the first
$N_{\mathrm{warm}}$ samples of free closing at the approach speed
(since free-running effort depends on jaw speed;
\hyperref[app:effort]{Appendix~\ref*{app:effort}}). After warm-up, the
detector compares the moving average of the last $N_{\mathrm{avg}}$
samples, $\bar{I}(t)$, with $I_0$ and declares contact when
\[
  \bar{I}(t) - I_0 \;\geq\; \Delta I_{\mathrm{th}}
\]
holds for $N_{\mathrm{persist}}$ consecutive ticks. Persistence prevents
false triggers under quantised, correlated noise (0 false latches in
$1500$ simulated empty closings), at the cost of a small pre-latch
compression that sets the detection floor (\S\ref{sec:floor}). We use
an amplitude threshold rather than a rate detector
($\mathrm{d}I/\mathrm{d}t$)~\cite{ma2020sensorless, kim2022collision}
because soft objects produce a shallow force ramp ($\dot{F} = k\dot{x}$)
and the servo load register is too noisy to differentiate reliably.

The object size is estimated as the jaw gap at the latch,
\[
  \hat{D} \;=\; g(x_c),
\]
where $g$ maps encoder position to jaw gap and is calibrated once per
gripper. An object resting off the fixed jaw is pushed before both jaws
engage, biasing $\hat{D}$ upward
(\hyperref[app:hardware]{Appendix~\ref*{app:hardware}}).

\subsection{Loading}
\label{sec:online-k}

Once the effort rises above baseline, the jaw slows to the fit speed
$v_{\mathrm{fit}}$, and after contact a proportional law drives it
toward the bound:
\[
  a(t) \;=\; \mathrm{clip}\bigl(k_p(\varepsilon - \hat{r}(t)),\;
     0,\; v_{\mathrm{fit}}\bigr).
\]
The first loading uses $\hat{k} = k_{\min}$, stopping at the floor
force $\varepsilon k_{\min}\hat{D}$. At each stop, the controller fits
the recorded pairs $(g(x), \hat{F})$ across all loadings with a shared
slope $s$ and one intercept per loading (to allow sub-millimetre
contact shifts). On the first loading, a two-piece linear fit (flat
followed by a line, with at least $n_{\min}$ samples per side) drops
the pre-contact approach and initial contact toe; subsequent loadings
start in contact and are used in full. To avoid over-compressing the
object, we take a lower confidence bound $s - z\sigma_s$ ($z = 2$),
with the standard error $\sigma_s$ inflated by
$\sqrt{(1 + \rho)/(1 - \rho)}$ for tick-to-tick noise correlation $\rho$.

Because the fit spans forces $[F_{\mathrm{low}}, F_{\mathrm{top}}]$,
its slope $s$ is a chord rather than the secant $F/\delta$ from zero
force. For a stiffening contact $F \propto \delta^{n}$ with
$r = F_{\mathrm{low}}/F_{\mathrm{top}}$ and exponent up to
$n_{\max} = 1.5$ (Hertz contact~\cite{johnson1985contact}), we convert
the chord to a conservative secant estimate:
\[
  \hat{k} \;=\; \max\!\left(k_{\min},\;
    (s - z\sigma_s)\,\frac{1 - r^{1/n_{\max}}}{1 - r}\right),
\]
which is exact for $n = n_{\max}$ and conservative for all
$n \le n_{\max}$. The floor $k_{\min}$ represents the secant stiffness
at $\varepsilon$ of the softest target fruit~\cite{sirisomboon2012evaluation,
ince2016effects}.

\subsection{Probing}
\label{sec:probe}

A single loading up to $\varepsilon k_{\min}\hat{D}$ spans only a few
effort quanta and rarely resolves the slope. Probing therefore climbs
from $k_{\min}$ in steps of
\[
  k_{i+1} \;=\; \min\bigl(\hat{k},\; (1 + s_{\max})\,k_i\bigr),
  \qquad k_0 = k_{\min},
\]
with $s_{\max} = 0.2$, in one of two variants that obey the same strain
bound (\S\ref{sec:alpha}; Table~\ref{tab:sim-release}):
\begin{itemize}
  \item \emph{Cyclic unload--reload} (backdrivable actuators): at each
    step $\varepsilon k_i\hat{D}$, the controller refits $\hat{k}$
    across all loadings so far, unloads at $v_{\mathrm{fit}}$ to
    $F_{\mathrm{low}} = 0.25\,\mathrm{N}$ (by at most $\Delta x_{\max}$,
    keeping the object against the fixed jaw), and reloads toward
    $\varepsilon k_{i+1}\hat{D}$.
  \item \emph{Monotonic stepping} (position-controlled servos that do
    not reverse under load): at each step $\varepsilon k_i\hat{D}$, the
    controller refits $\hat{k}$ over all post-contact samples so far and
    advances directly to $\varepsilon k_{i+1}\hat{D}$ along the same
    closing pass (\hyperref[app:hardware]{Appendix~\ref*{app:hardware}}).
\end{itemize}
Probing ends when $k_{i+1}$ would rise by less than $5\,\%$ and the
slope is resolved ($z\sigma_s \le 0.1\,s$), or after
$N_{\mathrm{probe}} = 8$ steps.

\begin{figure}[pos=htbp]
  \centering
  \includegraphics[width=0.7\columnwidth]{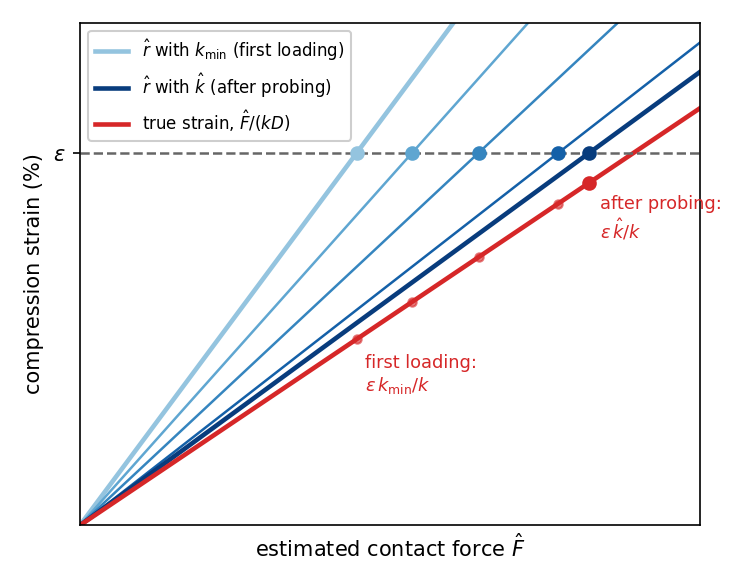}
  \caption{Strain bound (\S\ref{sec:alpha}). Each loading stops where
    its estimate $\hat{r}$ reaches $\varepsilon$ (blue dots). Its
    stiffness is a lower bound on the object's stiffness $k$, so the true
    strain there is $\varepsilon\,\hat{k}/k \le \varepsilon$ (red dots).
    The first loading uses the floor $k_{\min}$; probing raises
    $\hat{k}$ towards $k$, and the margin to $\varepsilon$ shrinks.}
  \label{fig:bound-schematic}
\end{figure}

\subsection{Hold}
\label{sec:stop-policy}

At the first tick with $\hat{r} \ge \varepsilon$ that does not start a
probe, the controller commits the grasp, records the jaw position
$x_h$, and holds it through the lift:
\[
  a(t) \;=\; \mathrm{clip}\bigl(k_h(x(t) - x_h),\; 0,\; v_{\mathrm{fit}}\bigr).
\]
We hold position rather than keep tracking $\hat{r}$ because motor
effort senses force on the moving jaw only: during the lift, gravity
shifts load between the jaws, so apparent effort can drop without any
loss of compression, which would cause a force-tracking law to close
further. Position hold also avoids creeping closed on noisy samples
near $\varepsilon$.

\subsection{Strain bound}
\label{sec:alpha}

Let $F$ be the true force at the stop, $\alpha$ the true effort-per-force
coefficient, $\tilde{\alpha}$ the calibrated value
($F = (\tilde{\alpha}/\alpha)\hat{F}$), and $k_s = F/\delta$ the
object's secant stiffness at the stop. Stopping at
$\hat{F} = \varepsilon\hat{k}\hat{D}$ (Fig.~\ref{fig:bound-schematic})
yields a true strain of
\[
  r_{\mathrm{true}} \;=\; \frac{F}{k_s D}
  \;=\; \varepsilon\,
        \frac{\hat{k}}{k_s}\,
        \frac{\hat{D}}{D}\,
        \frac{\tilde{\alpha}}{\alpha}.
\]
When $\hat{k}$ comes from the fit, it scales with $\alpha/\tilde{\alpha}$
and the last factor cancels; when $\hat{k} = k_{\min}$, the margin
$k_s/k_{\min}$ absorbs it.

Thus $r_{\mathrm{true}} \le \varepsilon$ holds when:
\begin{itemize}
  \item $k_s \ge k_{\min}$ and contact stiffens no faster than
    $F \propto \delta^{n_{\max}}$,
  \item $\hat{D} \le D$,
  \item $\varepsilon$ lies above the detection floor (\S\ref{sec:floor}),
  \item the pre-contact baseline $I_0$ remains valid at the loading
    speed (Table~\ref{tab:sim-noise}), and
  \item the lower confidence bound does not exceed the true slope
    ($\approx 98\,\%$ per single fit, slightly lower across up to eight
    probe refits).
\end{itemize}

\section{Experimental setup}
\label{sec:experiments}

\subsection{Simulation}
\label{sec:sim}

We simulate the SO-ARM101~\cite{soarm101} in MuJoCo~3.8.0~\cite{todorov2012mujoco}
with the gym-soarm model~\cite{gym_soarm} from
LeRobot~\cite{cadene2024lerobot}; after the stop the arm lifts the
object and carries it sideways. The object is a rigid cube of
$15$--$50\,\mathrm{mm}$ and $3$--$140\,\mathrm{g}$ whose compliance
sits in a spring behind each jaw pad, so the true strain is known.
The spring is a power law $F \propto \delta^{n}$ with secant stiffness
$k$ at $1\,\%$ strain; $k \in \{2, 4, 6, 8, 10\} \times 10^3\,\mathrm{N/m}$
spans tomato from overripe to firm~\cite{sirisomboon2012evaluation},
and $n = 1.5$, Hertz contact~\cite{johnson1985contact}, in the main
runs, with $n = 1$ and $1.25$ as checks of the estimator's
conservative side. The motor current is linear in the jaw force, with
free-running current, noise, drift and quantisation identified from
75 hardware trials.
\hyperref[app:sim]{Appendix~\ref*{app:sim}} lists all parameters.

\subsection{Hardware}
\label{sec:hardware}

Hardware trials use the SO-ARM101 with its Feetech STS3215 servos and
taped pads, with the gripper servo's load register as the force proxy.
The jaw-gap map, the effort coefficient $\alpha$ and the gripper's own
deflection, which is subtracted before the fit, are calibrated once
(\hyperref[app:hardware]{Appendix~\ref*{app:hardware}}). We grasp
20 cherry and 12 medium tomatoes
(Table~\ref{tab:fruit-results}), 5 cherry and 3 medium per setting,
each twice on faces $90^\circ$ apart ($n = 10$ and $6$); the two
grasps of a fruit are not independent, so we report counts per kind.
Each grasp lifts the fruit and carries it
to a box, and an operator, blind to the setting, scores the two
contact patches at release from 0 (none) to 3 (split).

\subsection{Baselines and ablations}
\label{sec:baselines}

\emph{Fixed force.} The default without force sensing, it shows what
one threshold can do across stiffness: after contact the jaw closes
until $\hat{F} \ge F_{\mathrm{th}}$. Fixed~$0\,\mathrm{N}$ stops at
detection, the gentlest stop the detector allows. In simulation
$F_{\mathrm{th}}$ is swept (Fig.~\ref{fig:sim-fth}) and the tables
report $0$, $2.5$ and $5\,\mathrm{N}$; on hardware
Fixed~$10\,\mathrm{N}$ adds a margin to the $8\,\mathrm{N}$ that
carried a medium tomato in the pilot.

\emph{Ablations.} To isolate each estimate, we replace $\hat{k}$ or
$\hat{D}$ by its true value or its prior; with both at their prior
the stop is a size-scaled threshold $\varepsilon k_{\min} D$, which
knows the size but not the stiffness. We also vary the number of probe
cycles from none to eight.

\subsection{Metrics and operating range}
\label{sec:metrics}

A trial \textbf{succeeds} if the cube remains with the gripper after
transport (\hyperref[app:metrics]{Appendix~\ref*{app:metrics}}), or if
the fruit is released inside the box. It counts as \textbf{damaged} if
simulated strain stays above $1.5\,\%$ for five consecutive ticks, between
cell deformation ($0.5$--$1.0\,\%$) and visible bruising
($3\,\%$) in tomato~\cite{li2017mathematical}, or if either contact patch
on hardware scores above zero.
In simulation, the main tables report trials within the gripper's
feasible \textbf{range}: $k \ge k_{\min}$, carryable within the bound
($m g / \mu \le \varepsilon k D$), and detectable within the bound
($F_{\mathrm{det}} \le \varepsilon k D$, with $F_{\mathrm{det}} = 1.2\,\mathrm{N}$
from \S\ref{sec:floor}). Out-of-range trials (including $D \in [50, 60]\,\mathrm{mm}$,
where jaw tilt causes slip) are reported in
\hyperref[app:limits]{Appendix~\ref*{app:limits}}.

\section{Results}
\label{sec:results}

In simulation we set $\varepsilon = 1.25\,\%$ and
$k_{\min} = 4000\,\mathrm{N/m}$ (medium-ripe
tomato~\cite{sirisomboon2012evaluation}). We run $100$ trials per
stiffness and method with randomised $D$, density and sensor noise;
$330$ of the $400$ trials with $k \ge k_{\min}$ fall within the
operating range (\S\ref{sec:metrics}).

\subsection{No fixed threshold both holds and protects}
\label{sec:results-fixed}
\begin{figure}[!htbp]
  \centering
  \includegraphics[width=0.85\columnwidth]{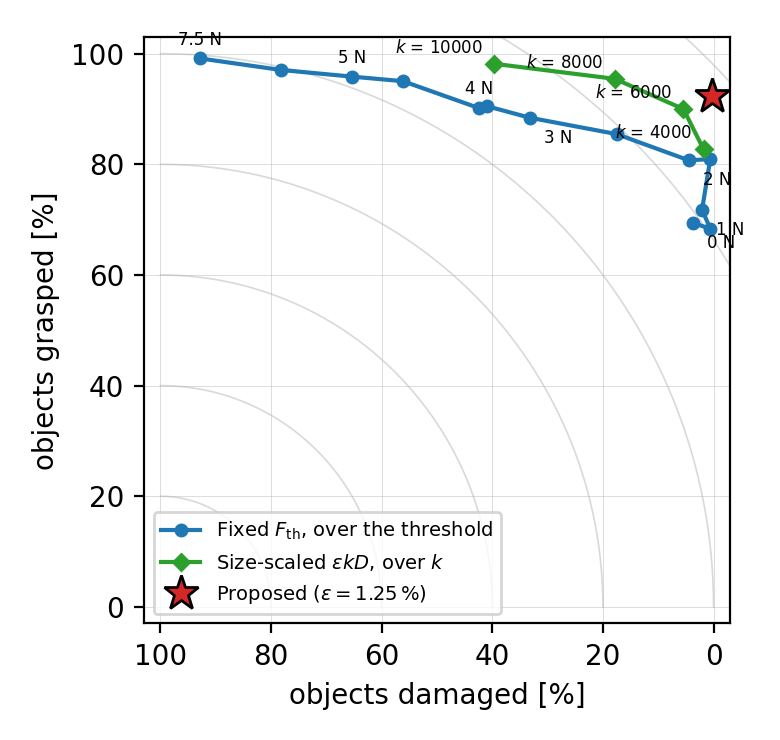}
  \caption{Grasp rate against damage rate over the range (damage
    decreasing to the right) for the fixed-force stop, the size-scaled
    stop, and the proposed stop. Rings show equal-weight distance from
    the bottom-left corner.}
  \label{fig:sim-fth}
\end{figure}

Table~\ref{tab:sim-main} and Fig.~\ref{fig:sim-fth} compare the
controllers across stiffness and object size. Fixed~$0\,\mathrm{N}$
avoids damage ($4\,\%$) but drops nearly all $40$--$50\,\mathrm{mm}$
objects, whereas Fixed~$5\,\mathrm{N}$ carries $94\,\%$ of objects but
damages $65\,\%$. Scaling a fixed stiffness with size
($\varepsilon k D$) shows the same trade-off. The proposed controller
carries $92\,\%$ of objects ($89$--$95\,\%$) with $0.3\,\%$ damage
($0$--$2\,\%$); the true strain at the stop exceeded $\varepsilon$ in
$3\,\%$ of trials (by at most $16\,\%$), and remained below the
$1.5\,\%$ damage line throughout the lift.

\begin{table*}[t]
  \centering
  \caption{Grasp success\,/\,damage rate (\%) per stiffness and per
    object size within the method's range, $D$ uniform in
    $15$--$50\,\mathrm{mm}$, $100$ trials per stiffness and method
    before the range is applied, rounded to whole per cent.}
  \label{tab:sim-main}
  \small
  \begin{tabular}{lcccccccc}
\toprule
\multirow{2}{*}{Method} & \multicolumn{4}{c}{$k$ (N/m)} & \multicolumn{3}{c}{$D$ (mm)} & \multirow{2}{*}{Average} \\
\cmidrule(lr){2-5}\cmidrule(lr){6-8}
 & 4000 & 6000 & 8000 & 10\,000 & 15--25 & 25--40 & 40--50 & \\
\midrule
Fixed $0\,\mathrm{N}$ & 84 / 7 & 72 / 5 & 66 / 4 & 64 / 1 & 100 / 2 & 82 / 6 & 5 / 1 & 69 / 4 \\
Fixed $2.5\,\mathrm{N}$ & 100 / 52 & 83 / 26 & 77 / 10 & 84 / 0 & 100 / 40 & 100 / 13 & 34 / 0 & 84 / 17 \\
Fixed $5\,\mathrm{N}$ & 100 / 100 & 99 / 95 & 87 / 57 & 94 / 30 & 100 / 100 & 100 / 65 & 71 / 24 & 94 / 65 \\
\midrule
Proposed ($\varepsilon = 1.25\,\%$) & 100 / 2 & 92 / 0 & 91 / 0 & 91 / 0 & 100 / 0 & 100 / 1 & 68 / 0 & 92 / 0 \\
\bottomrule
\end{tabular}

\end{table*}

\begin{figure}[!htbp]
  \centering
  \includegraphics[width=0.85\columnwidth]{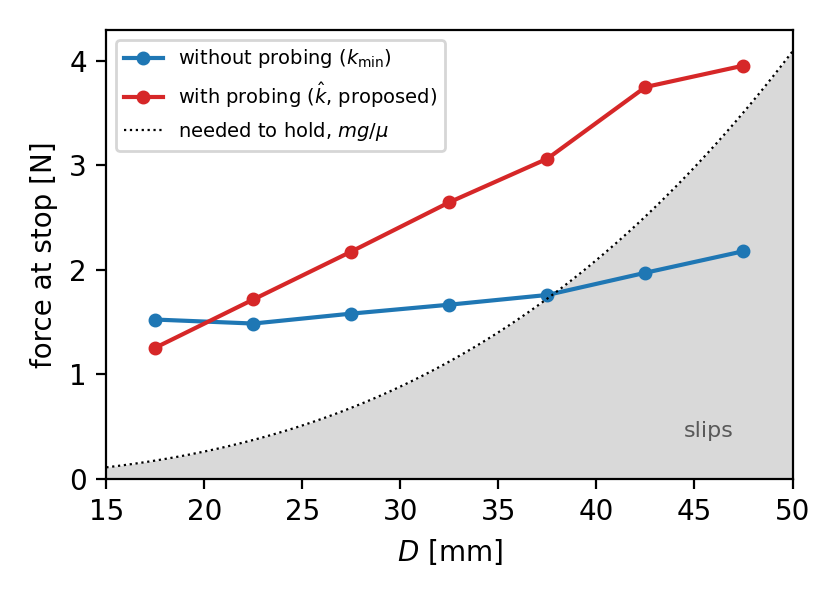}
  \caption{Force at the stop with and without probing over object
    size, on objects above the floor ($k \ge 6000\,\mathrm{N/m}$).
    Below the dotted line, $m g / \mu$, the object slips.}
  \label{fig:sim-ablation}
\end{figure}

\subsection{Probing carries the heavy objects}
\label{sec:results-estimate}
\begin{figure}[!htbp]
  \centering
  \includegraphics[width=0.85\columnwidth]{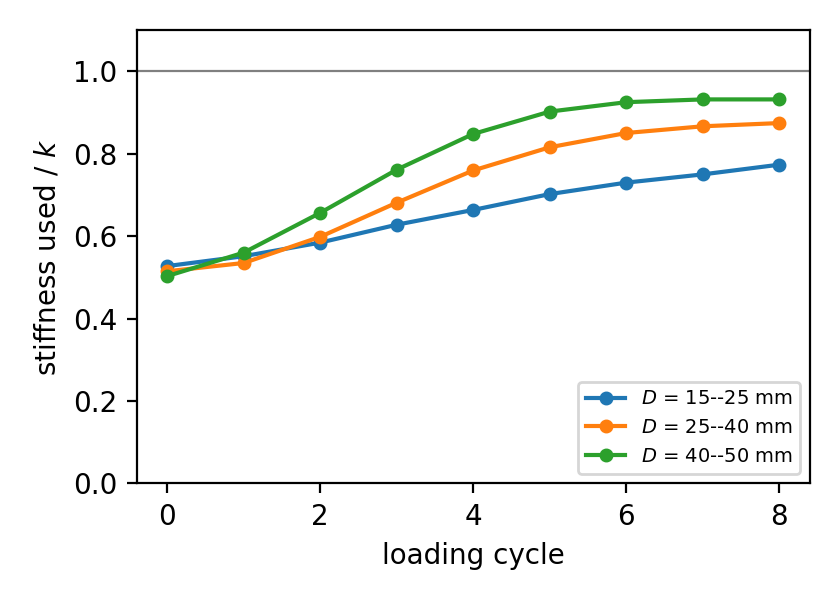}
  \caption{Stiffness used by each loading of the adaptive stop, over
    the true stiffness, per object size on objects above the floor
    ($k > k_{\min}$). Loading $0$ uses $k_{\min}$; a trial that has
    stopped keeps its last value.}
  \label{fig:sim-probing}
\end{figure}
Without probing, the stop force $\varepsilon k_{\min} \hat{D}$ scales
with $D$ while object weight scales with $D^3$. Above $40\,\mathrm{mm}$
the floor force falls below $m g / \mu$ (Fig.~\ref{fig:sim-ablation}),
carrying only $28\,\%$ of $40$--$50\,\mathrm{mm}$ objects
(Table~\ref{tab:sim-ablation}).

Probing raises the stiffness used toward the true $k$ over successive
loadings (Fig.~\ref{fig:sim-probing}). The overall grasp rate rises from
$84\,\%$ after the first loading to $94\,\%$ after five, and from
$28\,\%$ to $73\,\%$ on $40$--$50\,\mathrm{mm}$ objects, while damage
stays at or below $0.3\,\%$ (Table~\ref{tab:sim-probing}). Of the
$50\,\mathrm{s}$ total closing time, $40\,\mathrm{s}$ is free-air
approach before contact ($N = 0$) and adaptive probing itself takes
$10\,\mathrm{s}$ ($\approx 1.5\,\mathrm{s}$ per cycle). Replacing
$\hat{D}$ with the true size changes the average rates by at most one
percentage point (\hyperref[app:results]{Appendix~\ref*{app:results}}).

\subsection{Limits and robustness}
\label{sec:results-limits}
\begin{figure}[!htbp]
  \centering
  \includegraphics[width=0.85\columnwidth]{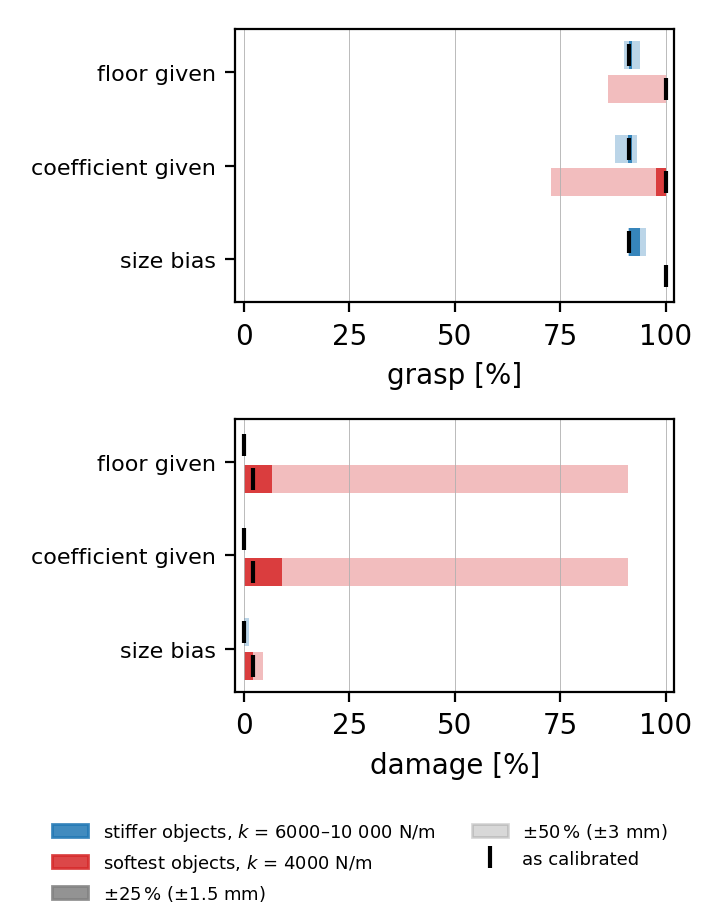}
  \caption{Grasp and damage rates over the range under $\pm 25\,\%$
    (dark) and $\pm 50\,\%$ (light) errors in $k_{\min}$ and $\alpha$,
    and $\pm 1.5$ / $\pm 3\,\mathrm{mm}$ bias in $\hat{D}$ (ticks mark
    nominal values; blue: $k > k_{\min}$, red: $k = k_{\min}$).}
  \label{fig:sim-sensitivity}
\end{figure}

\label{sec:results-sensitivity}
Outside the operating range (\hyperref[app:limits]{Appendix~\ref*{app:limits}}),
the proposed controller drops objects that are too heavy without
damaging them ($0\,\%$ damage), whereas objects softer than $k_{\min}$
are over-compressed already on the first loading.

Fig.~\ref{fig:sim-sensitivity} shows sensitivity to errors in $k_{\min}$,
$\alpha$, and $\hat{D}$. Above the floor ($k > k_{\min}$), damage stays
below $1\,\%$ because $\hat{k}$ replaces $k_{\min}$ and $\alpha$ cancels
(\S\ref{sec:alpha}). At the floor ($k = k_{\min}$), overestimating
$k_{\min}$ or $\alpha$ by $1.5\times$ raises damage to $91\,\%$. Among
the sensor-noise terms, the controller is most sensitive to a
speed-dependent step in free-running current ($39\,\%$ damage at
$+0.5\,\mathrm{N}$; \hyperref[app:robustness]{Appendix~\ref*{app:robustness}}).

\subsection{The detection floor}
\label{sec:floor}
\begin{figure}[!htbp]
  \centering
  \includegraphics[width=0.85\columnwidth]{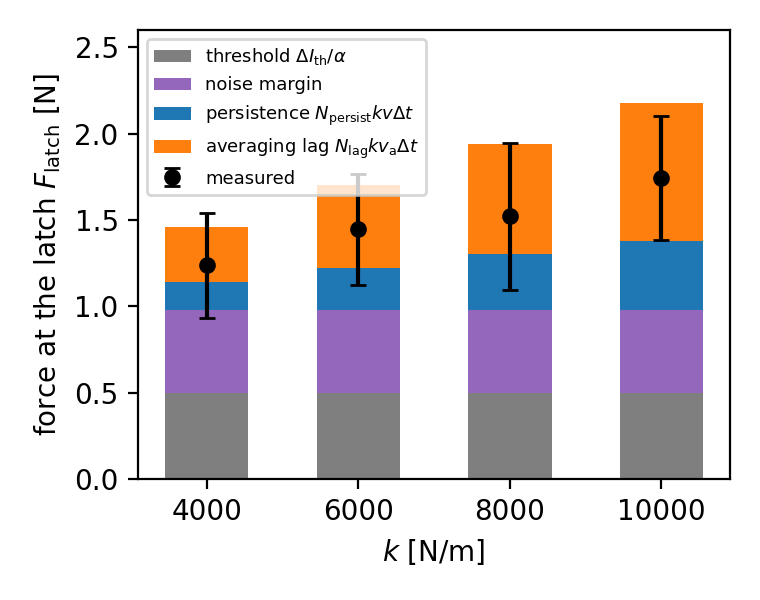}
  \caption{Force on the object when contact is declared per stiffness
    at $v = 0.2\,\mathrm{mm/s}$, showing the four stacked bound terms
    and the measured mean and s.d.}
  \label{fig:sim-floor}
\end{figure}

Because the jaw keeps closing while contact is confirmed, the force at
declaration is bounded by
\[
  F_{\mathrm{latch}} \;\lesssim\; \frac{\Delta I_{\mathrm{th}}}{\alpha}
      \;+\; F_{\sigma}
      \;+\; N_{\mathrm{persist}}\, k\, v\, \Delta t
      \;+\; N_{\mathrm{lag}}\, k\, v_{\mathrm{a}}\, \Delta t,
\]
combining the threshold, noise margin ($F_{\sigma} = 0.5\,\mathrm{N}$),
and jaw travel during persistence ($v$) and averaging lag
($v_{\mathrm{a}}$, $N_{\mathrm{lag}} = 4$). In simulation
$F_{\mathrm{latch}}$ is $1.2$--$1.7\,\mathrm{N}$, staying
$0.2$--$0.4\,\mathrm{N}$ below this bound (Fig.~\ref{fig:sim-floor}).
Dividing by $k D$ gives the minimum strain spent on detection: at
$k_{\min}$ it reaches $\varepsilon = 1.25\,\%$ below $D = 25\,\mathrm{mm}$,
setting the small-object limit of the range. Raising the loading speed
to $v = 1\,\mathrm{mm/s}$ increases $F_{\mathrm{latch}}$ to
$4.5\,\mathrm{N}$ and damages $38\,\%$ of objects
(Table~\ref{tab:sim-speed}).

\subsection{Fruit}
\label{sec:results-fruit}

\begin{figure}[!htbp]
  \centering
  \includegraphics[width=0.85\columnwidth]{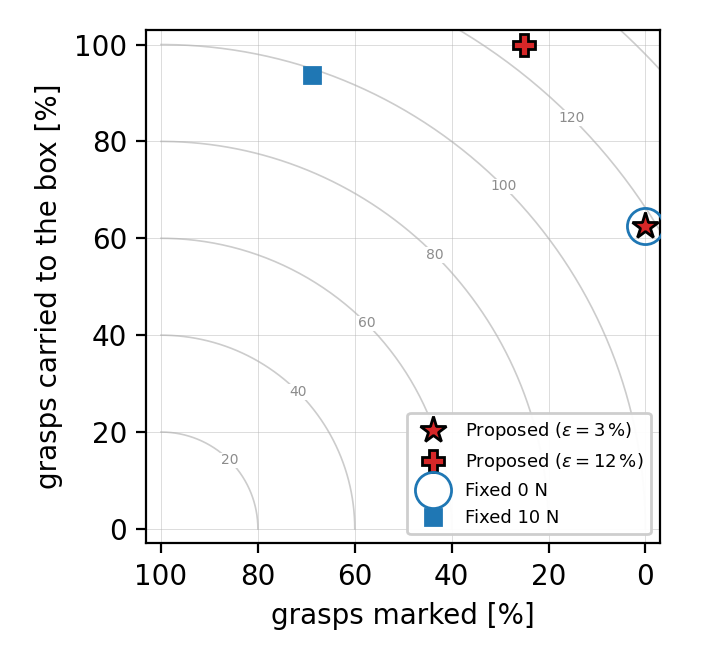}
  \caption{Share of grasps carried to the box against share marked
    (decreasing to the right) over 16 grasps per setting (10 cherry,
    6 medium tomatoes).}
  \label{fig:fruit-pareto}
\end{figure}

\begin{table*}[t]
  \centering
  \caption{Tomato trials on the SO-ARM101, per kind of tomato: grasps
    carried to the box and grasps marked out of $n$ ($n = 10$ on cherry
    and $6$ on medium tomatoes), two grasps per fruit on faces
    $90^\circ$ apart and one setting per fruit. The stop force,
    $\hat{k}$ and the contact strain at the stop are medians (range);
    the strain is the stop force over the median fitted contact
    stiffness of the kind and the diameter. $m g / \mu$ is the force
    the fruit needs to be carried, with $\mu = 0.3$.}
  \label{tab:fruit-results}
  \small
  \begin{tabular}{lccccc}
    \toprule
    Setting & Stop force [N] & $\hat{k}$ [N/m] & Contact strain (\%) & Carried & Marked \\
    \midrule
    \multicolumn{6}{l}{\emph{Cherry tomato}: $D$ 26--30\,mm, 9--15\,g, $m g / \mu$ 0.3--0.5\,N} \\
\quad Proposed ($\varepsilon = 3\,\%$) & 0.8 (0.6--1.7) & 500 (500--1248) & 2 (2--5) & 10/10 & 0/10 \\
\quad Proposed ($\varepsilon = 12\,\%$) & 3.8 (1.6--6.0) & 1100 (500--1529) & 11 (5--17) & 10/10 & 1/10 \\
\quad Fixed $0\,\mathrm{N}$ & 0.9 (0.6--1.2) & -- & 3 (2--3) & 10/10 & 0/10 \\
\quad Fixed $10\,\mathrm{N}$ & 10.0 & -- & 30 (27--31) & 10/10 & 8/10 \\
\midrule
\multicolumn{6}{l}{\emph{Medium tomato}: $D$ 59--69\,mm, 109--167\,g, $m g / \mu$ 3.6--5.5\,N} \\
\quad Proposed ($\varepsilon = 3\,\%$) & 1.4 (1.0--3.3) & 897 (500--1496) & 2 (1--3) & 0/6 & 0/6 \\
\quad Proposed ($\varepsilon = 12\,\%$) & 12.2 (7.5--13.3) & 1537 (1035--1652) & 13 (8--15) & 6/6 & 3/6 \\
\quad Fixed $0\,\mathrm{N}$ & 1.1 (0.9--1.4) & -- & 1 (1--1) & 0/6 & 0/6 \\
\quad Fixed $10\,\mathrm{N}$ & 10.0 & -- & 10 (10--11) & 5/6 & 3/6 \\
 
    \bottomrule
  \end{tabular}
\end{table*}

\begin{figure*}[pos=tp]
  \centering
  \includegraphics[width=0.9\textwidth]{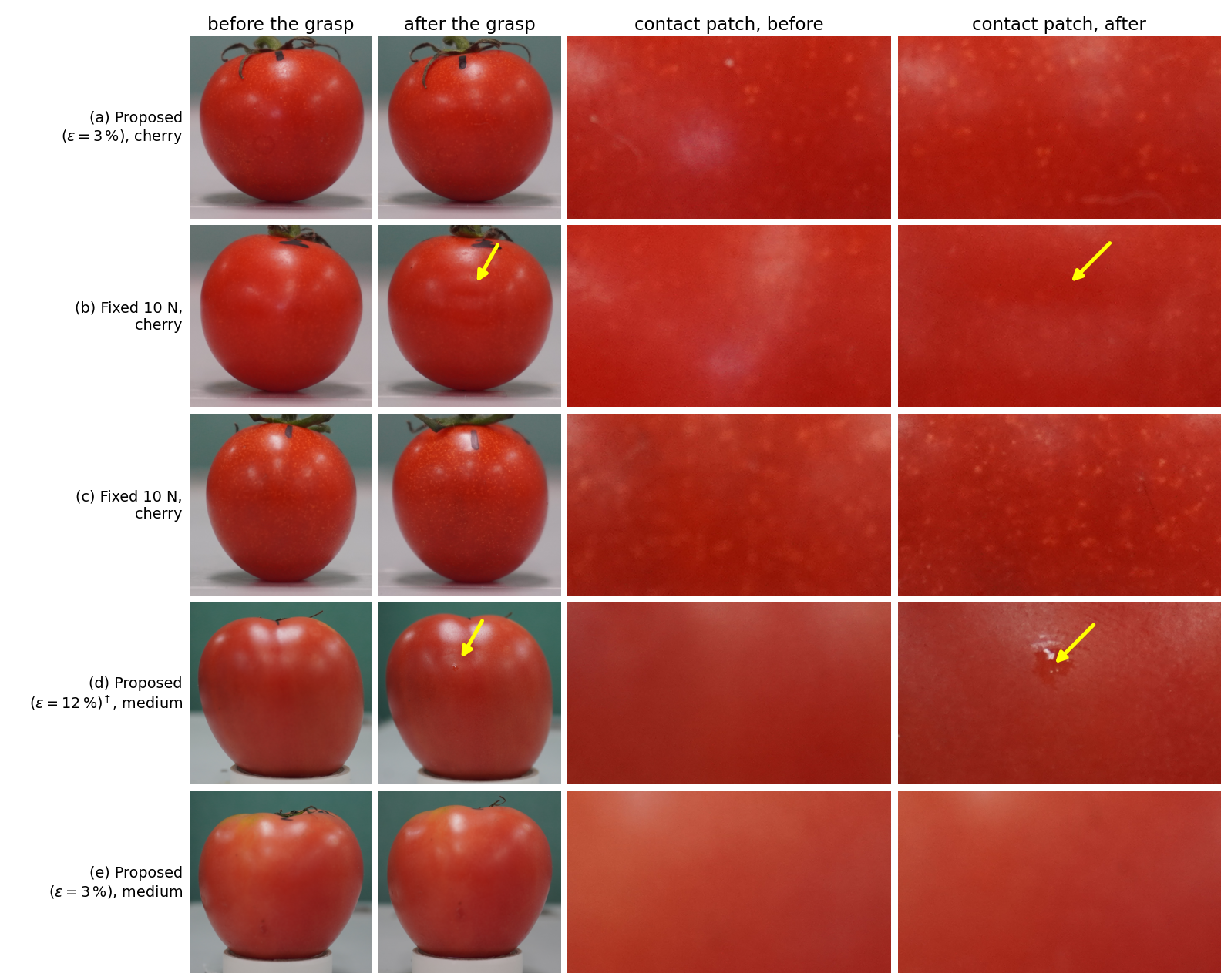}
  \caption{Example grasps, photographed from the front before and after
    the grasp, with the contact patch enlarged; arrows point at visible
    marks. (a) Proposed ($\varepsilon = 3\,\%$) on a cherry tomato,
    $0.9\,\mathrm{N}$: carried, no mark. (b) Fixed~$10\,\mathrm{N}$ on a
    cherry tomato: carried, a dent along the pad. (c) Fixed~$10\,\mathrm{N}$
    on another cherry tomato: carried and scored as marked although the
    patch looks unchanged; it was soft to the touch. (d) Proposed
    ($\varepsilon = 12\,\%$) on a medium tomato,
    $13\,\mathrm{N}$: carried, a small mark at the contact. (e) Proposed
    ($\varepsilon = 3\,\%$) on a medium tomato, $3.3\,\mathrm{N}$: dropped, no mark.}
  \label{fig:fruit-photos}
\end{figure*}

On tomatoes, the controller bounds contact indentation
$\hat{F}/(\hat{k}\hat{D})$, where the fitted pad-contact stiffness
($\hat{k} = 710$--$1650\,\mathrm{N/m}$, floor
$k_{\min} = 500\,\mathrm{N/m}$ from pilot grasps) is several times
lower than whole-fruit plate stiffness
($2000$--$10\,000\,\mathrm{N/m}$~\cite{sirisomboon2012evaluation})
because the taped pads indent a curved surface locally. We tested both
the literature's $3\,\%$ whole-fruit damage
threshold~\cite{li2017mathematical} ($\varepsilon = 3\,\%$) and
$\varepsilon = 12\,\%$, chosen to place medium tomatoes within the
carry range ($m g / \mu \le \varepsilon \hat{k} D$ requires
$\varepsilon \ge 9\,\%$ at the median $\hat{k}$, plus a margin).

\emph{Carry, contact strain, and marks (Fig.~\ref{fig:fruit-pareto}, Table~\ref{tab:fruit-results}, Fig.~\ref{fig:fruit-photos}).}
Settings stopping at $\le 3\,\%$ computed contact strain
(Fixed~$0\,\mathrm{N}$ and Proposed at $\varepsilon = 3\,\%$) left 0
marks in 32 grasps and carried all cherry tomatoes, but their stop
forces ($0.9$--$1.4\,\mathrm{N}$) fell below $m g / \mu$
($3.6$--$5.5\,\mathrm{N}$) on medium tomatoes, carrying none.
Fixed~$10\,\mathrm{N}$ carried 15 of 16 grasps but compressed cherry
tomatoes to $30\,\%$ strain, marking 8 of 10 cherry grasps (11 of 16
overall). Proposed ($\varepsilon = 12\,\%$) carried all 16 grasps at
$11$--$13\,\%$ median strain, reducing cherry markings to 1 of 10 while
marking 3 of 6 on medium tomatoes (matching Fixed~$10\,\mathrm{N}$).

\emph{Force adaptation and timing.}
At $\varepsilon = 3\,\%$, contact detection alone ($\approx 1\,\mathrm{N}$)
exceeded the floor target on cherry tomatoes, leaving $\hat{k}$ at
$k_{\min}$ in 11 of 16 grasps. At $\varepsilon = 12\,\%$, $\hat{k}$
rose above $k_{\min}$ in 13 of 16 grasps and the stop force tracked
$\varepsilon \hat{k} \hat{D}$ with a median ratio of $1.03$ (at most
$1.22$), scaling threefold from $3.8\,\mathrm{N}$ on cherry to
$12.2\,\mathrm{N}$ on medium tomatoes. Across all 32 proposed grasps,
$F_{\mathrm{stop}} \ge m g / \mu$ separated carried from dropped trials
without error (Table~\ref{tab:fruit-results}). Closing at
$\varepsilon = 12\,\%$ took a median of $30\,\mathrm{s}$ ($22\,\mathrm{s}$
free-air approach and $10\,\mathrm{s}$ post-contact probing:
$7\,\mathrm{s}$ on cherry, $22\,\mathrm{s}$ on medium tomatoes).

\section{Discussion}
\label{sec:discussion}

\subsection{An intuitive bound in place of a force sensor}
The results show that a strain bound, combined with an estimate of
the stiffness, sets a suitable grip force without a force or tactile
sensor. The operator enters two values once per crop and gripper, the
strain the fruit tolerates, $\varepsilon$, and the stiffness floor
$k_{\min}$; neither depends on the fruit at hand. Whole-fruit tests
report both~\cite{li2017mathematical, sirisomboon2012evaluation}, but
at the jaw contact the floor had to be set lower (\S\ref{sec:future}).
No fixed force sets a suitable grip across sizes and
stiffnesses (\S\ref{sec:results-fixed}), whereas the bound with the
estimate carries $92\,\%$ of the objects in its range with $0.3\,\%$
damage. The estimate is what makes the bound usable: with the floor
alone the stop drops the heavy objects, and probing raises the force
only as far as the data support (\S\ref{sec:results-estimate}).

\subsection{The bound as a parameter for learned grasping}
Because $\varepsilon$ is intuitive for a person, it may also suit a
learned policy. A handling policy would otherwise have to learn
damage avoidance from reward and risk crushing fruit while it
explores. With this controller as its grasp primitive, the policy
chooses only $\varepsilon$, and any $\varepsilon$ between the
detection floor and the fruit's damage strain keeps the compression
bounded under the conditions of \S\ref{sec:alpha}, much as a shield does in safe reinforcement
learning~\cite{alshiekh2018safe}. Controller-parameter action spaces
are known to ease learning in contact-rich
manipulation~\cite{martinmartin2019variable}.

\subsection{What the fruit trials showed}
\label{sec:future}
While the simulation evaluates rates on a contact model parameterized
from whole-fruit literature, the hardware trials test the stop
mechanism on real tomatoes and expose three physical factors outside
the controller:

\emph{Contact versus whole-fruit strain.}
Because $\hat{k}$ reflects local curved-surface and pad compliance
rather than bulk plate compression, the computed contact strain
$\hat{F}/(\hat{k}\hat{D})$ is larger than plate strain at the same
force, making the $3\,\%$ plate threshold conservative at the pad
contact (marks appeared only from $\approx 10\,\%$ computed strain).
Calibrating $\varepsilon$ and $k_{\min}$ directly from a
pad-indentation test on the target crop would align both parameters
with the gripper's contact geometry.

\emph{Pad friction.}
With $\mu = 0.3$, holding a $170\,\mathrm{g}$ medium tomato during
transport requires about $8\,\mathrm{N}$, or $8$--$10\,\%$ contact strain,
close to where marks appeared. Higher-friction pads
or gentler transport accelerations would lower the required carry force
without changing the controller.

\emph{Detection floor on hardware.}
Because the servo closed three times faster in jaw gap than the
simulation, its detection force ($\approx 1\,\mathrm{N}$) exceeded the
$\varepsilon = 3\,\%$ floor target on cherry tomatoes before probing
could begin. Slower closing near contact or a quieter effort signal
would lower this floor.

\subsection{Limitations}
\label{sec:limitations}
Several limitations remain. First, contact detection compresses the
object before control begins, setting a minimum attainable strain that
rises with closing speed (\S\ref{sec:floor}). Second, blind closing from a wide-open jaw takes $30$--$50\,\mathrm{s}$
(Table~\ref{tab:sim-probing}), though $70$--$80\,\%$ of this time is
free-air approach ($40\,\mathrm{s}$ in simulation, $10$--$22\,\mathrm{s}$
on hardware) while post-contact probing takes $\approx 10\,\mathrm{s}$;
pre-shaping the jaw with vision or proximity sensing would remove the
approach time, and inferring stiffness from the contact
transient~\cite{kutsuzawa2024learning} could shorten probing further.
Third, objects softer than $k_{\min}$ are over-compressed, and objects
at $k_{\min}$ absorb any calibration error in $k_{\min}$, $\alpha$, or
speed-dependent current offsets (\S\ref{sec:results-limits},
Table~\ref{tab:sim-noise}). On hardware, sample sizes were modest
($16$ grasps per setting across two tomato varieties), gripper
deflection had to be subtracted, and viscoelastic creep, ripeness
grading, and detachment forces during harvesting~\cite{elfferich2022soft}
were not addressed.

\section{Conclusion}
\label{sec:conclusion}

We presented a sensorless controller that bounds compression strain
using only the encoder and motor effort of a servo gripper. Given a
strain bound $\varepsilon$ and a stiffness floor $k_{\min}$ set once
per crop and gripper, the controller estimates contact size at
detection, refines stiffness conservatively by probing, and stops at
$\varepsilon \hat{k} \hat{D}$. In simulation, where no fixed force
threshold both holds heavy objects and protects soft ones, the proposed
stop carries $92\,\%$ of objects within its operating range with
$0.3\,\%$ damage. On real cherry and medium tomatoes, the stop force
tracked $\varepsilon \hat{k} \hat{D}$ when set above the detection
floor ($\varepsilon = 12\,\%$), adapting grip force threefold across
fruit sizes and reducing markings on cherry tomatoes from $8/10$ (under
a fixed $10\,\mathrm{N}$ threshold) to $1/10$ while carrying all
fruits; on medium tomatoes both marked $3/6$.

\section*{Data availability}
The simulation environment, controller implementation, hardware
driver, and experiment data are available at
\url{https://github.com/shutouyusei/damage-safe-grasping}.

\section*{Declaration of generative AI and AI-assisted technologies in the writing process}
During the preparation of this work the authors used Claude
(Anthropic) in order to improve the language and readability of the
manuscript. After using this tool, the authors reviewed and edited the
content as needed and take full responsibility for the content of the
publication.

\onecolumn
\appendix
\setcounter{table}{0}
\renewcommand{\thetable}{\thesection.\arabic{table}}
\setcounter{figure}{0}
\renewcommand{\thefigure}{\thesection.\arabic{figure}}
\setcounter{topnumber}{4}\setcounter{bottomnumber}{4}\setcounter{totalnumber}{8}
\renewcommand{\topfraction}{0.95}\renewcommand{\bottomfraction}{0.95}\renewcommand{\textfraction}{0.05}
\setlength{\intextsep}{8pt plus 2pt minus 2pt}
\setlength{\floatsep}{8pt plus 2pt minus 2pt}
\setlength{\textfloatsep}{10pt plus 2pt minus 2pt}
\stretchtables

\section{Simulation and hardware details}
\label{app:sim}

\subsection{Setup and controller settings}
\label{app:contact}
\label{app:settings}
\label{app:metrics}

Each jaw pad is mounted on a critically damped normal spring with rigid
pad--cube contact so that true compression is known
(Table~\ref{tab:sim-params}). Table~\ref{tab:settings} lists the
controller parameters and metric thresholds.

\begin{table}[pos=!ht]
  \centering
  \caption{Simulation parameters.}
  \label{tab:sim-params}
  \small\setlength{\tabcolsep}{4pt}
  \begin{tabular}{ll}
    \toprule
    Parameter & Value \\
    \midrule
    Physics step; control rate & $0.5\,\mathrm{ms}$; $50\,\mathrm{Hz}$ \\
    Approach; loading speed $v_{\mathrm{fit}}$ & $1\,\mathrm{mm/s}$; $0.2\,\mathrm{mm/s}$ \\
    Cube side $D$; density & uniform in $15$--$50\,\mathrm{mm}$; $900$--$1100\,\mathrm{kg/m^3}$ \\
    Friction (pad--cube; cube--table) & $0.3$; $0.05$ \\
    Pad spring & $F = k\delta_{\mathrm{ref}}(\delta/\delta_{\mathrm{ref}})^{n}$ ($\delta_{\mathrm{ref}}$ at $1\,\%$ strain), critically damped~\cite{hunt1975coefficient} \\
    Stiffness $k$; exponent $n$ & $\{2, 4, 6, 8, 10\} \times 10^3\,\mathrm{N/m}$; $1.5$ (main), $1.0$, $1.25$ \\
    Pad--cube contact & time constant $2\,\mathrm{ms}$, impedance $0.99$ ($< 0.02\,\mathrm{mm/N}$) \\
    Arm servos; lift and carry & STS3215 ($3.35\,\mathrm{N\,m}$ limit); $30\,\mathrm{mm}$ lift ($5\,\mathrm{s}$), $150\,\mathrm{mm}$ carry ($1\,\mathrm{s}$) \\
    \bottomrule
  \end{tabular}
\end{table}

\begin{table}[pos=!ht]
  \centering
  \caption{Controller settings and metric thresholds.}
  \label{tab:settings}
  \small\setlength{\tabcolsep}{4pt}
  \begin{tabular}{ll}
    \toprule
    Setting & Value \\
    \midrule
    Detector $\Delta I_{\mathrm{th}}$; $N_{\mathrm{warm}}$; $N_{\mathrm{avg}}$; $N_{\mathrm{persist}}$ & $0.01\,\mathrm{A}$ ($0.5\,\mathrm{N}$); $50$; $3$; $10$ \\
    Nominal $\alpha$; gains $k_p$, $k_h$ & $0.02\,\mathrm{A/N}$; $2\,/\,\%$ strain, $2\,/\,\mathrm{mm}$ drift (normalised to $5\,\mathrm{mm/s}$) \\
    Breakpoint fit $n_{\min}$; s.e.\ inflation & $4$; $1.36$ ($\rho = 0.3$) \\
    Probe backoff $\Delta x_{\max}$; $N_{\mathrm{probe}}$ & $2\,\mathrm{mm}$; $8$ \\
    Size map $g$; fixed-force close & 23 cubes ($15$--$70\,\mathrm{mm}$, error $< 0.4\,\mathrm{mm}$); $k_F = 125\,\mathrm{A}^{-1}$ \\
    \midrule
    Grasp success & rise $\ge 50\,\%$ of gripper; gap drift $\le 20\,\mathrm{mm}$; speed $< 0.05\,\mathrm{m/s}$ \\
    Strain at stop; carried strain & median of ticks 11--20 after commit; median at top of lift \\
    Timeout & $120\,\mathrm{s}$ ($400\,\mathrm{s}$ for proposed); none occurred \\
    \bottomrule
  \end{tabular}
\end{table}

\subsection{Effort signal}
\label{app:effort}

We model the motor current at time $t$ as
$I(t) = Q_q\bigl(I^{\ast}(t)\,(1 + \sigma_r\,\nu(t)) + d\,t\bigr)$ with
$I^{\ast}(t) = I_f(v(t)) + \alpha F_m(t)$, moving-jaw force $F_m$,
speed-dependent free-running current
$I_f(v)$~\cite{linderoth2013robotic}, first-order autoregressive noise
$\nu$ (coefficient $\rho$, unit variance), drift $d$, and quantisation
$Q_q$ of step $q$ (Table~\ref{tab:noise}).

\begin{table}[pos=!ht]
  \centering
  \caption{Effort-noise parameters identified from 75 hardware trials ($15.4$ units/N) and values used in the noise sweep.}
  \label{tab:noise}
  \small\setlength{\tabcolsep}{4pt}
  \begin{tabular}{llccc}
    \toprule
    Symbol & Meaning & Hardware logs & Main runs & Noise sweep \\
    \midrule
    $I_{\mathrm{idle}}$ & free-running current & $\approx 60$ units ($4\,\mathrm{N}$) & $0.08\,\mathrm{A}$ & -- \\
    $\Delta I_f$ (load) & step at $v_{\mathrm{fit}}$ & $\pm 8$ units & $0$ & $\pm 0.01, \pm 0.02\,\mathrm{A}$ \\
    $\Delta I_f$ (open) & step while opening & $+15$ units & $0$ & -- \\
    $\sigma_r$ & relative noise s.d. & $\approx 7\,\%$ & $0.07$ & $0$--$0.21$ \\
    $\rho$ & lag-1 autocorrelation & $0.1$--$0.4$ & $0.3$ & $0, 0.6, 0.8$ \\
    $q$ & quantisation step & $4$ units ($0.27\,\mathrm{N}$) & $0.0053\,\mathrm{A}$ & $0, 2\times, 3\times$ \\
    $d$ & current drift & -- & $0$ & $\pm 0.1, \pm 0.2\,\mathrm{mA/s}$ \\
    \bottomrule
  \end{tabular}
\end{table}

\subsection{Hardware calibration and protocol}
\label{app:hardware}

Table~\ref{tab:hw-params} summarises the hardware calibration and
protocol. The Feetech servo uses the monotonic-stepping variant
(\S\ref{sec:probe}), fitting all samples since contact once the jaw has
moved. To guard against stick--slip of a few encoder counts, the fit
bounds compliance $\mathrm{d}\delta/\mathrm{d}F$ from above by its
standard error plus one encoder count before inverting for $\hat{k}$
(without this margin, half of simulated stops exceed $\varepsilon$;
Table~\ref{tab:sim-release}). Thus $\hat{k}$ rises above $k_{\min}$
only when compression at $\varepsilon$ spans many encoder counts
($26$--$70$ on tomatoes at $\varepsilon = 12\,\%$, versus $2$--$5$ in
simulation).

\begin{table}[pos=!ht]
  \centering
  \caption{Hardware calibration and protocol.}
  \label{tab:hw-params}
  \small\setlength{\tabcolsep}{4pt}
  \begin{tabular}{ll}
    \toprule
    Item & Value \\
    \midrule
    Rate; speeds & $50\,\mathrm{Hz}$; $3\,\mathrm{mm/s}$ approach, $0.6\,\mathrm{mm/s}$ loading in jaw gap ($F_{\mathrm{det}} \approx 1\,\mathrm{N}$) \\
    Contact detector & settings of Table~\ref{tab:settings} (0 false latches in 20 free closings) \\
    Jaw-gap map $g$ & 7 blocks ($20$--$80\,\mathrm{mm}$); $\hat{D} - D$: $-2.2$ to $+0.5\,\mathrm{mm}$ (cherry), $+0.25$ to $+4.4\,\mathrm{mm}$ (medium) \\
    Effort coeff.\ $\alpha$ & $15.4$ units/N on scale ($+4 \pm 16\,\%$ over 15 closes at $1.25$--$5\,\mathrm{N}$) \\
    Gripper deflection & $2000$--$3700\,\mathrm{N/m}$ (rigid blocks of $30, 40, 50\,\mathrm{mm}$), subtracted before fit \\
    Encoder step & $0.12\,\mathrm{mm}$ jaw closure per count \\
    Protocol & fruit against fixed jaw at equator; close, lift, transport to box, release \\
    Damage score & 0 none, 1 indent or softening to touch, 2 permanent set, 3 split \\
    \bottomrule
  \end{tabular}
\end{table}

\FloatBarrier
\section{Additional results}
\label{app:results}
\setcounter{table}{0}
\setcounter{figure}{0}

\subsection{Stop rule and probing}
\label{app:ablation}

Table~\ref{tab:sim-ablation} replaces each estimated input of
$\hat{F} = \varepsilon \hat{k}\hat{D}$ with its prior or true value,
Table~\ref{tab:sim-probing} varies the number of probe cycles $N$,
and Table~\ref{tab:sim-release} compares the two probing variants
(\S\ref{sec:probe}).

\begin{table}[pos=!ht]
  \centering
  \caption{Grasp\,/\,damage rate (\%) per stiffness and per object size
    over the range for the stop rule $\hat{F} = \varepsilon k D$ with
    each estimated input replaced by its prior or its true value.}
  \label{tab:sim-ablation}
  \small
  \begin{tabular}{lcccccccc}
\toprule
\multirow{2}{*}{Stop rule} & \multicolumn{4}{c}{$k$ (N/m)} & \multicolumn{3}{c}{$D$ (mm)} & \multirow{2}{*}{Average} \\
\cmidrule(lr){2-5}\cmidrule(lr){6-8}
 & 4000 & 6000 & 8000 & 10\,000 & 15--25 & 25--40 & 40--50 & \\
\midrule
$k_{\min}$, true $D$ & 100 / 5 & 86 / 2 & 76 / 2 & 79 / 0 & 100 / 2 & 97 / 3 & 31 / 0 & 83 / 2 \\
$k_{\min}$, $\hat{D}$ & 100 / 0 & 90 / 1 & 78 / 0 & 76 / 0 & 100 / 0 & 99 / 0 & 28 / 1 & 84 / 0 \\
$\hat{k}$, true $D$ & 100 / 0 & 94 / 0 & 88 / 0 & 93 / 0 & 100 / 0 & 100 / 0 & 68 / 0 & 93 / 0 \\
$\hat{k}$, $\hat{D}$ (proposed) & 100 / 2 & 92 / 0 & 91 / 0 & 91 / 0 & 100 / 0 & 100 / 1 & 68 / 0 & 92 / 0 \\
\bottomrule
\end{tabular}

\end{table}

\begin{table}[pos=!ht]
  \begin{minipage}[t]{0.52\linewidth}
    \centering\setcaptionwidth{\linewidth}
    \caption{Fixed $N$ probe cycles over the range (shares in \%;
      $k_s$: secant stiffness at stop).}
    \label{tab:sim-probing}
    \footnotesize\setlength{\tabcolsep}{3pt}
    \begin{tabular}{lccccc}
\toprule
Cycles $N$ & $\hat{k}/k_s$ & $\hat{k} > k_s$ & $r > \varepsilon$ & Grasp & Time (s) \\
\midrule
0 & 0.71 & 7 & 4 & 84 & 40 \\
1 & 0.78 & 5 & 1 & 85 & 41 \\
2 & 0.81 & 5 & 1 & 89 & 43 \\
3 & 0.84 & 6 & 1 & 90 & 44 \\
5 & 0.87 & 6 & 2 & 94 & 47 \\
8 & 0.90 & 10 & 2 & 95 & 52 \\
Adaptive (6.8) & 0.89 & 7 & 3 & 92 & 50 \\
\bottomrule
\end{tabular}

  \end{minipage}\hfill
  \begin{minipage}[t]{0.44\linewidth}
    \centering\setcaptionwidth{\linewidth}
    \caption{Rates (\%) over the range for the two probing variants
      (\S\ref{sec:probe}, \hyperref[app:hardware]{Appendix~\ref*{app:hardware}}).}
    \label{tab:sim-release}
    \footnotesize\setlength{\tabcolsep}{3pt}
    \begin{tabular}{lcccc}
\toprule
Probing variant & Grasp & Damage & $r > \varepsilon$ & $\bar{r}/\varepsilon$ \\
\midrule
Cyclic unload--reload & 92 & 0 & 3 & 0.81 \\
Monotonic stepping & 89 & 1 & 6 & 0.72 \\
\bottomrule
\end{tabular}

  \end{minipage}
\end{table}

\subsection{Outside the range}
\label{app:limits}

Table~\ref{tab:sim-limits} evaluates the controllers of
Table~\ref{tab:sim-main} on trials violating each feasibility condition
(\S\ref{sec:metrics}). Above $D = 50\,\mathrm{mm}$, objects slip out
during transport at any force because the rotating jaw pads, parallel at
$40\,\mathrm{mm}$, tilt by $8^\circ$ at $50\,\mathrm{mm}$ and $17^\circ$
at $60\,\mathrm{mm}$.

\begin{table}[pos=!ht]
  \centering
  \caption{Grasp\,/\,damage rate (\%) outside the range per violated
    condition (\S\ref{sec:metrics}), with trial counts in parentheses.}
  \label{tab:sim-limits}
  \small
  \begin{tabular}{lcccc}
\toprule
Method & $k < k_{\min}$ & $m g / \mu > \varepsilon k D$ & $F_{\mathrm{det}} > \varepsilon k D$ & $D$ = 50--60\,mm \\
\midrule
Fixed $0\,\mathrm{N}$ & 57 / 42 (100) & 8 / 0 (39) & 100 / 35 (31) & 0 / 0 (400) \\
Fixed $2.5\,\mathrm{N}$ & 86 / 100 (100) & 50 / 8 (36) & 100 / 100 (26) & 0 / 0 (400) \\
Fixed $5\,\mathrm{N}$ & 100 / 100 (100) & 78 / 94 (32) & 100 / 100 (31) & 0 / 23 (400) \\
\midrule
Proposed ($\varepsilon = 1.25\,\%$) & 84 / 100 (100) & 36 / 0 (39) & 100 / 26 (31) & 0 / 0 (400) \\
\bottomrule
\end{tabular}

\end{table}

\FloatBarrier
\subsection{Sensitivity and robustness}
\label{app:robustness}

Tables~\ref{tab:sim-sensitivity}--\ref{tab:sim-speed} detail parameter
sensitivity, estimator accuracy across contact exponents $n$, effort-noise
sweeps, and loading-speed sweeps over the operating range. In
Table~\ref{tab:sim-noise}, downward current drift causes force
underestimation ($16\,\%$ damage at $-0.2\,\mathrm{mA/s}$), upward
drift weakens the grip ($43\,\%$ grasp at $+0.2\,\mathrm{mA/s}$), and a
$+0.01\,\mathrm{A}$ ($0.5\,\mathrm{N}$) step at the loading speed
triggers early contact detection ($39\,\%$ damage).

\begin{table}[pos=!ht]
  \begin{minipage}[t]{0.485\linewidth}
    \centering\setcaptionwidth{\linewidth}
    \caption{Rates (\%) over the range under miscalibrated $k_{\min}$,
      $\alpha$, or $\hat{D}$, and alternative bounds $\varepsilon$.}
    \label{tab:sim-sensitivity}
    \footnotesize\setlength{\tabcolsep}{3.5pt}
    \begin{tabular}{lcccc}
\toprule
Variant & Grasp & Damage & $r > \varepsilon$ & $\bar{r}/\varepsilon$ \\
\midrule
Nominal ($\varepsilon = 1.25\,\%$) & 92 & 0 & 3 & 0.81 \\
$\tilde{\alpha}/\alpha = 0.5$ / $1.5$ & 86 / 94 & 0 / 12 & 0 / 16 & 0.67 / 0.89 \\
$\hat{D} - D = -3$ / $+3\,\mathrm{mm}$ & 92 / 96 & 0 / 2 & 0 / 13 & 0.74 / 0.90 \\
$k_{\min}\;\times 0.5$ / $\times 1.5$ & 90 / 95 & 0 / 12 & 0 / 16 & 0.73 / 0.89 \\
$\varepsilon = 0.7\,\%$ / $2.0\,\%$ & 57 / 100 & 2 / 97 & 18 / 1 & 0.87 / 0.79 \\
\bottomrule
\end{tabular}

  \end{minipage}\hfill
  \begin{minipage}[t]{0.485\linewidth}
    \centering\setcaptionwidth{\linewidth}
    \caption{Estimator accuracy on linear ($n = 1$) and stiffening
      objects over the range (shares in \%).}
    \label{tab:sim-estimation}
    \footnotesize\setlength{\tabcolsep}{3.5pt}
    \begin{tabular}{lccccc}
\toprule
Object & $\hat{k}/k_s$ & $\hat{k} > k_s$ & $r > \varepsilon$ & Mean $r/\varepsilon$ & Max $r/\varepsilon$ \\
\midrule
$n = 1.0$ & 0.80 & 1 & 3 & 0.75 & 1.15 \\
$n = 1.25$ & 0.84 & 3 & 2 & 0.78 & 1.13 \\
$n = 1.5$ & 0.89 & 7 & 3 & 0.81 & 1.16 \\
\bottomrule
\end{tabular}

  \end{minipage}
\end{table}

\begin{table}[pos=!ht]
  \begin{minipage}[t]{0.485\linewidth}
    \centering\setcaptionwidth{\linewidth}
    \caption{Rates (\%) over the range with one effort-noise parameter
      varied at a time (Table~\ref{tab:noise}).}
    \label{tab:sim-noise}
    \footnotesize\setlength{\tabcolsep}{3.5pt}
    \begin{tabular}{lcccc}
\toprule
Variant & Grasp & Damage & $r > \varepsilon$ & $\bar{r}/\varepsilon$ \\
\midrule
Nominal & 92 & 0 & 3 & 0.81 \\
$d = -0.2$ / $+0.2\,\mathrm{mA/s}$ & 98 / 43 & 16 / 1 & 43 / 0 & 1.00 / 0.35 \\
$q = 0$ / $3\times$ & 93 / 93 & 0 / 0 & 3 / 2 & 0.81 / 0.81 \\
$\sigma_r = 0$ / $0.21$ & 95 / 88 & 0 / 48 & 1 / 48 & 0.85 / 1.00 \\
$\rho = 0$ / $0.8$ & 93 / 97 & 0 / 35 & 1 / 43 & 0.82 / 1.10 \\
$\Delta I_f = -0.01$ / $+0.01\,\mathrm{A}$ & 95 / 96 & 5 / 39 & 18 / 43 & 0.90 / 1.06 \\
\bottomrule
\end{tabular}

  \end{minipage}\hfill
  \begin{minipage}[t]{0.485\linewidth}
    \centering\setcaptionwidth{\linewidth}
    \caption{Rates (\%) over the range against loading speed $v$
      ($\varepsilon = 1.25\,\%$).}
    \label{tab:sim-speed}
    \footnotesize\setlength{\tabcolsep}{3.5pt}
    \begin{tabular}{lcccc}
\toprule
Loading speed $v$ & Grasp & Damage & $r > \varepsilon$ & $\bar{r}/\varepsilon$ \\
\midrule
$0.05\,\mathrm{mm/s}$ & 91 & 0 & 0 & 0.73 \\
$0.1\,\mathrm{mm/s}$ & 93 & 0 & 1 & 0.79 \\
$0.2\,\mathrm{mm/s}$ (nominal) & 92 & 0 & 3 & 0.81 \\
$0.5\,\mathrm{mm/s}$ & 94 & 6 & 7 & 0.84 \\
$1.0\,\mathrm{mm/s}$ & 93 & 38 & 14 & 0.85 \\
\bottomrule
\end{tabular}

  \end{minipage}
\end{table}
\FloatBarrier

\clearpage
\twocolumn
\bibliographystyle{unsrtnat}
\bibliography{refs}

\begin{thebibliography}{37}
\providecommand{\natexlab}[1]{#1}
\providecommand{\url}[1]{\texttt{#1}}
\expandafter\ifx\csname urlstyle\endcsname\relax
  \providecommand{\doi}[1]{doi: #1}\else
  \providecommand{\doi}{doi: \begingroup \urlstyle{rm}\Url}\fi

\bibitem[Chauhan et~al.(2022)Chauhan, Brouwer, and Westra]{chauhan2022robotics}
Aneesh Chauhan, Bastiaan Brouwer, and Eelke Westra.
\newblock Robotics for a quality-driven post-harvest supply chain.
\newblock \emph{Current Robotics Reports}, 3\penalty0 (2):\penalty0 39--48,
  2022.
\newblock \doi{10.1007/s43154-022-00075-8}.

\bibitem[Wang et~al.(2025)Wang, Tu, Xu, Zhang, Knoll, Zhou, and
  Ying]{wang2025towards}
Qingyu Wang, Yuyang Tu, Weidong Xu, Jianwei Zhang, Alois Knoll, Mingchuan Zhou,
  and Yibin Ying.
\newblock Towards damage-less robotic fragile fruit grasping: A systematic
  review on system design, end effector, and visual and tactile feedback.
\newblock \emph{Journal of Field Robotics}, 42\penalty0 (8):\penalty0
  4521--4543, 2025.

\bibitem[Sirisomboon et~al.(2012)Sirisomboon, Tanaka, and
  Kojima]{sirisomboon2012evaluation}
Panmanas Sirisomboon, Munehiro Tanaka, and Takayuki Kojima.
\newblock Evaluation of tomato textural mechanical properties.
\newblock \emph{Journal of Food Engineering}, 111:\penalty0 618--624, 2012.

\bibitem[Wang et~al.(2026)Wang, Zhang, Knoll, Cui, Jiang, and
  Zhou]{wang2026adaptive}
Qingyu Wang, Jianwei Zhang, Alois Knoll, Di~Cui, Huanyu Jiang, and Mingchuan
  Zhou.
\newblock Towards damage-less robotic fragile fruit grasping: Learning adaptive
  grasping force from multimodal data and human demonstration.
\newblock \emph{Computers and Electronics in Agriculture}, 240:\penalty0
  111194, 2026.
\newblock \doi{10.1016/j.compag.2025.111194}.

\bibitem[Yu et~al.(2025)Yu, Ji, Zhang, Ruan, Xu, and Wu]{yu2025grasping}
Xiaowei Yu, Wei Ji, Hongwei Zhang, Chengzhi Ruan, Bo~Xu, and Kaiyang Wu.
\newblock Grasping force optimization and {DDPG} impedance control for apple
  picking robot end-effector.
\newblock \emph{Agriculture}, 15\penalty0 (10):\penalty0 1018, 2025.
\newblock \doi{10.3390/agriculture15101018}.

\bibitem[Li et~al.(2017)Li, Andrews, and Wang]{li2017mathematical}
Zhiguo Li, James Andrews, and Yuqing Wang.
\newblock Mathematical modelling of mechanical damage to tomato fruits.
\newblock \emph{Postharvest Biology and Technology}, 126:\penalty0 50--56,
  2017.

\bibitem[Li et~al.(2013)Li, Li, Yang, and Liu]{li2013internal}
Zhiguo Li, Pingping Li, Hongling Yang, and Jizhan Liu.
\newblock Internal mechanical damage prediction in tomato compression using
  multiscale finite element models.
\newblock \emph{Journal of Food Engineering}, 116\penalty0 (3):\penalty0
  639--647, 2013.
\newblock \doi{10.1016/j.jfoodeng.2013.01.016}.

\bibitem[Yuan et~al.(2017)Yuan, Dong, and Adelson]{yuan2017gelsight}
Wenzhen Yuan, Siyuan Dong, and Edward~H Adelson.
\newblock Gelsight: High-resolution robot tactile sensors for estimating
  geometry and force.
\newblock \emph{Sensors}, 17\penalty0 (12):\penalty0 2762, 2017.

\bibitem[Han et~al.(2024)Han, Yu, Batra, Boyd, Mehta, Zhao, She, Hutchinson,
  and Zhao]{han2024learning}
Yunhai Han, Kelin Yu, Rahul Batra, Nathan Boyd, Chaitanya Mehta, Tuo Zhao,
  Yu~She, Seth Hutchinson, and Ye~Zhao.
\newblock Learning generalizable vision-tactile robotic grasping strategy for
  deformable objects via transformer.
\newblock \emph{IEEE/ASME Transactions on Mechatronics}, 30\penalty0
  (1):\penalty0 554--566, 2024.

\bibitem[Giannico et~al.(2017)Giannico, Castaman, and
  Ghidoni]{giannico2017evaluation}
Simone Giannico, Nicola Castaman, and Stefano Ghidoni.
\newblock Evaluation of the force-current relationship in a 3-finger
  underactuated gripper.
\newblock In \emph{2017 European Conference on Mobile Robots (ECMR)}, pages
  1--6. IEEE, 2017.

\bibitem[Ballesteros et~al.(2020)Ballesteros, Pastor, G{\'o}mez-de Gabriel,
  Gandarias, Garc{\'i}a-Cerezo, and Urdiales]{ballesteros2020proprioceptive}
Joaquin Ballesteros, Francisco Pastor, Jes{\'u}s~M G{\'o}mez-de Gabriel, Juan~M
  Gandarias, Alfonso~J Garc{\'i}a-Cerezo, and Cristina Urdiales.
\newblock Proprioceptive estimation of forces using underactuated fingers for
  robot-initiated {pHRI}.
\newblock \emph{Sensors}, 20\penalty0 (10):\penalty0 2863, 2020.

\bibitem[Opara and Pathare(2014)]{opara2014bruise}
Umezuruike~Linus Opara and Pankaj~B Pathare.
\newblock Bruise damage measurement and analysis of fresh horticultural
  produce—a review.
\newblock \emph{Postharvest Biology and Technology}, 91:\penalty0 9--24, 2014.

\bibitem[Al-Dairi et~al.(2022)Al-Dairi, Pathare, Al-Yahyai, and
  Opara]{al2022mechanical}
Mai Al-Dairi, Pankaj~B Pathare, Rashid Al-Yahyai, and Umezuruike~Linus Opara.
\newblock Mechanical damage of fresh produce in postharvest transportation:
  Current status and future prospects.
\newblock \emph{Trends in Food Science \& Technology}, 124:\penalty0 195--207,
  2022.

\bibitem[Zhang et~al.(2024)Zhang, Lammers, Chu, Li, and Lu]{zhang2024automated}
Kaixiang Zhang, Kyle Lammers, Pengyu Chu, Zhaojian Li, and Renfu Lu.
\newblock An automated apple harvesting robot---from system design to field
  evaluation.
\newblock \emph{Journal of Field Robotics}, 41\penalty0 (7):\penalty0
  2384--2400, 2024.

\bibitem[Xiong et~al.(2020)Xiong, Ge, Grimstad, and From]{xiong2020autonomous}
Ya~Xiong, Yuanyue Ge, Lars Grimstad, and P{\aa}l~J From.
\newblock An autonomous strawberry-harvesting robot: Design, development,
  integration, and field evaluation.
\newblock \emph{Journal of Field Robotics}, 37\penalty0 (2):\penalty0 202--224,
  2020.

\bibitem[Lehnert et~al.(2017)Lehnert, English, McCool, Tow, and
  Perez]{lehnert2017autonomous}
Christopher Lehnert, Andrew English, Christopher McCool, Adam~W Tow, and
  Tristan Perez.
\newblock Autonomous sweet pepper harvesting for protected cropping systems.
\newblock \emph{IEEE Robotics and Automation Letters}, 2\penalty0 (2):\penalty0
  872--879, 2017.

\bibitem[Wang et~al.(2023)Wang, Kang, Zhou, Au, Wang, and Chen]{wang2023soft}
Xing Wang, Hanwen Kang, Hongyu Zhou, Wesley Au, Michael~Yu Wang, and Chao Chen.
\newblock Development and evaluation of a robust soft robotic gripper for apple
  harvesting.
\newblock \emph{Computers and Electronics in Agriculture}, 204:\penalty0
  107552, 2023.

\bibitem[Elfferich et~al.(2022)Elfferich, Dodou, and
  Della~Santina]{elfferich2022soft}
Johannes~F. Elfferich, Dimitra Dodou, and Cosimo Della~Santina.
\newblock Soft robotic grippers for crop handling or harvesting: A review.
\newblock \emph{IEEE Access}, 10:\penalty0 75428--75443, 2022.
\newblock \doi{10.1109/ACCESS.2022.3190863}.

\bibitem[Liu et~al.(2024)Liu, Zhang, Lou, Zhang, Zhou, and Chen]{liu2024soft}
Yuchen Liu, Jintao Zhang, Yuanxin Lou, Baohua Zhang, Jun Zhou, and Jiajie Chen.
\newblock Soft bionic gripper with tactile sensing and slip detection for
  damage-free grasping of fragile fruits and vegetables.
\newblock \emph{Computers and Electronics in Agriculture}, 220:\penalty0
  108904, 2024.

\bibitem[Cui et~al.(2020)Cui, Wang, Wei, Li, and Wang]{cui2020grasp}
Shaowei Cui, Rui Wang, Junhang Wei, Fanrong Li, and Shuo Wang.
\newblock Grasp state assessment of deformable objects using visual-tactile
  fusion perception.
\newblock In \emph{2020 IEEE International Conference on Robotics and
  Automation (ICRA)}, pages 538--544. IEEE, 2020.

\bibitem[De~Luca et~al.(2006)De~Luca, Albu-Schaffer, Haddadin, and
  Hirzinger]{deluca2006collision}
Alessandro De~Luca, Alin Albu-Schaffer, Sami Haddadin, and Gerd Hirzinger.
\newblock Collision detection and safe reaction with the {DLR}-{III}
  lightweight manipulator arm.
\newblock In \emph{2006 IEEE/RSJ International Conference on Intelligent Robots
  and Systems}, pages 1623--1630. IEEE, 2006.

\bibitem[Liu et~al.(2021)Liu, Wang, and Wang]{liu2021sensorless}
Sichao Liu, Lihui Wang, and Xi~Vincent Wang.
\newblock Sensorless force estimation for industrial robots using disturbance
  observer and neural learning of friction approximation.
\newblock \emph{Robotics and Computer-Integrated Manufacturing}, 71:\penalty0
  102168, 2021.

\bibitem[Wu et~al.(2025)Wu, Dong, Li, Bao, Dong, and Sun]{wu2025contact}
Peizhang Wu, Hui Dong, Pengfei Li, Yifei Bao, Wei Dong, and Lining Sun.
\newblock A new contact force estimation method for heavy robots without force
  sensors by combining {CNN-GRU} and force transformation.
\newblock \emph{Technologies}, 13\penalty0 (5):\penalty0 192, 2025.

\bibitem[Yu et~al.(2023)Yu, Huang, Zhao, Zhou, and Ou]{yu2023defgrasp}
Xinyi Yu, Rui Huang, Chongliang Zhao, Libo Zhou, and Linlin Ou.
\newblock Def-grasp: A robot grasping detection method for deformable objects
  without force sensor.
\newblock \emph{Neural Processing Letters}, 55\penalty0 (8):\penalty0
  11739--11756, 2023.

\bibitem[Kutsuzawa et~al.(2024)Kutsuzawa, Matsumoto, Owaki, and
  Hayashibe]{kutsuzawa2024learning}
Kyo Kutsuzawa, Minami Matsumoto, Dai Owaki, and Mitsuhiro Hayashibe.
\newblock Learning-based object's stiffness and shape estimation with
  confidence level in multi-fingered hand grasping.
\newblock \emph{Frontiers in Neurorobotics}, 18:\penalty0 1466630, 2024.

\bibitem[Ma et~al.(2020)Ma, Dong, Han, Yan, and Zhou]{ma2020sensorless}
Linjie Ma, Longlei Dong, Yi~Han, Jian Yan, and Jiaming Zhou.
\newblock A method of sensorless collision detection based on motor current for
  robot manipulator.
\newblock \emph{International Journal of Applied Electromagnetics and
  Mechanics}, 64\penalty0 (1-4):\penalty0 237--244, 2020.
\newblock \doi{10.3233/jae-209327}.

\bibitem[Kim(2022)]{kim2022collision}
Joonyoung Kim.
\newblock Collision detection and reaction for a collaborative robot with
  sensorless admittance control.
\newblock \emph{Mechatronics}, 84:\penalty0 102811, 2022.
\newblock \doi{10.1016/j.mechatronics.2022.102811}.

\bibitem[Johnson(1985)]{johnson1985contact}
Kenneth~L. Johnson.
\newblock \emph{Contact Mechanics}.
\newblock Cambridge University Press, Cambridge, 1985.
\newblock \doi{10.1017/CBO9781139171731}.

\bibitem[Ince et~al.(2016)Ince, Cevik, and Vursavus]{ince2016effects}
Ahmet Ince, Melih~Yavuz Cevik, and Kubilay~Kazim Vursavus.
\newblock Effects of maturity stages on textural mechanical properties of
  tomato.
\newblock \emph{International Journal of Agricultural and Biological
  Engineering}, 9\penalty0 (6):\penalty0 200--206, 2016.
\newblock \doi{10.3965/j.ijabe.20160906.1999}.

\bibitem[Knight et~al.(2024)Knight, Kooijmans, Cadene, Alibert, Aractingi,
  Aubakirova, Zouitine, Martino, Palma, Pascal, and Wolf]{soarm101}
Rob Knight, Pepijn Kooijmans, R{\'e}mi Cadene, Simon Alibert, Michel Aractingi,
  Dana Aubakirova, Adil Zouitine, Russi Martino, Steven Palma, Caroline Pascal,
  and Thomas Wolf.
\newblock Standard open {SO-100} \& {SO-101} arms.
\newblock GitHub, \url{https://github.com/TheRobotStudio/SO-ARM100}, 2024.

\bibitem[Todorov et~al.(2012)Todorov, Erez, and Tassa]{todorov2012mujoco}
Emanuel Todorov, Tom Erez, and Yuval Tassa.
\newblock Mujoco: A physics engine for model-based control.
\newblock In \emph{2012 IEEE/RSJ International Conference on Intelligent Robots
  and Systems}, pages 5026--5033. IEEE, 2012.
\newblock \doi{10.1109/IROS.2012.6386109}.

\bibitem[{SO-ARM Development Team}(2024)]{gym_soarm}
{SO-ARM Development Team}.
\newblock Gym {SO-ARM}: A gymnasium environment for {SO-ARM101} manipulation.
\newblock \url{https://github.com/masato-ka/gym-soarm}, 2024.
\newblock Version 0.1.0.

\bibitem[Cadene et~al.(2024)Cadene, Alibert, Soare, Gallouedec, Zouitine,
  Palma, Kooijmans, Aractingi, Shukor, Aubakirova, Russi, Capuano, Pascal,
  Choghari, Moss, and Wolf]{cadene2024lerobot}
Remi Cadene, Simon Alibert, Alexander Soare, Quentin Gallouedec, Adil Zouitine,
  Steven Palma, Pepijn Kooijmans, Michel Aractingi, Mustafa Shukor, Dana
  Aubakirova, Martino Russi, Francesco Capuano, Caroline Pascal, Jade Choghari,
  Jess Moss, and Thomas Wolf.
\newblock Lerobot: State-of-the-art machine learning for real-world robotics in
  pytorch.
\newblock \url{https://github.com/huggingface/lerobot}, 2024.

\bibitem[Alshiekh et~al.(2018)Alshiekh, Bloem, Ehlers, K{\"o}nighofer, Niekum,
  and Topcu]{alshiekh2018safe}
Mohammed Alshiekh, Roderick Bloem, R{\"u}diger Ehlers, Bettina K{\"o}nighofer,
  Scott Niekum, and Ufuk Topcu.
\newblock Safe reinforcement learning via shielding.
\newblock In \emph{Proceedings of the AAAI Conference on Artificial
  Intelligence}, volume~32, pages 2669--2678, 2018.
\newblock \doi{10.1609/aaai.v32i1.11797}.

\bibitem[Mart{\'\i}n-Mart{\'\i}n et~al.(2019)Mart{\'\i}n-Mart{\'\i}n, Lee,
  Gardner, Savarese, Bohg, and Garg]{martinmartin2019variable}
Roberto Mart{\'\i}n-Mart{\'\i}n, Michelle~A. Lee, Rachel Gardner, Silvio
  Savarese, Jeannette Bohg, and Animesh Garg.
\newblock Variable impedance control in end-effector space: An action space for
  reinforcement learning in contact-rich tasks.
\newblock In \emph{2019 IEEE/RSJ International Conference on Intelligent Robots
  and Systems}, pages 1010--1017. IEEE, 2019.
\newblock \doi{10.1109/IROS40897.2019.8968201}.

\bibitem[Hunt and Crossley(1975)]{hunt1975coefficient}
K.~H. Hunt and F.~R.~E. Crossley.
\newblock Coefficient of restitution interpreted as damping in vibroimpact.
\newblock \emph{Journal of Applied Mechanics}, 42\penalty0 (2):\penalty0
  440--445, 1975.
\newblock \doi{10.1115/1.3423596}.

\bibitem[Linderoth et~al.(2013)Linderoth, Stolt, Robertsson, and
  Johansson]{linderoth2013robotic}
Magnus Linderoth, Andreas Stolt, Anders Robertsson, and Rolf Johansson.
\newblock Robotic force estimation using motor torques and modeling of low
  velocity friction disturbances.
\newblock In \emph{2013 IEEE/RSJ International Conference on Intelligent Robots
  and Systems}, pages 3550--3556, 2013.
\newblock \doi{10.1109/IROS.2013.6696862}.

\end{thebibliography}

\end{document}